\documentclass{article} 
\usepackage[T1]{fontenc}
\usepackage{iclr2027_conference,times}

\usepackage{amsmath,amsfonts,bm}

\def\eqref#1{equation~\ref{#1}}

\def\1{\bm{1}}

\DeclareMathAlphabet{\mathsfit}{\encodingdefault}{\sfdefault}{m}{sl}
\SetMathAlphabet{\mathsfit}{bold}{\encodingdefault}{\sfdefault}{bx}{n}

\usepackage{hyperref}
\usepackage{url}
\usepackage{graphicx}
\usepackage{wrapfig}
\usepackage{booktabs}
\usepackage{multirow}
\usepackage{xcolor}
\usepackage[section]{placeins}
\usepackage[capitalize,nameinlink,noabbrev]{cleveref}
\usepackage[labelfont=bf]{caption}

\definecolor{linkblue}{RGB}{0,60,120}

\hypersetup{
    colorlinks=true,
    linkcolor=linkblue,  
    citecolor=linkblue,  
    urlcolor=linkblue    
}

\usepackage{xspace}
\newcommand{\MD}{\texttt{MIDAS}\xspace}
\newcommand{\LD}{\texttt{LIDAS}\xspace}

\newcommand{\x}{{\mkern-1.5mu\times\mkern-1.5mu}}
\newcommand{\dash}{{\mkern2mu\text{-}\mkern2mu}}

\DeclareMathOperator{\LoopOp}{Loop}
\newcommand{\LoopModel}[1]{\ensuremath{\LoopOp\,#1}}

\newcommand{\LoopThreeSixty}{\LoopModel{4\dash12\x2\dash4}\xspace}
\newcommand{\LoopSeventeen}{\LoopModel{4\dash8\x2\dash4}\xspace}
\newcommand{\LoopThree}{\LoopModel{10\dash8\x2\dash10}\xspace}
\newcommand{\LoopTwoSix}{\LoopModel{2\x6}\xspace}
\newcommand{\LoopFourThree}{\LoopModel{4\x3}\xspace}
\newcommand{\LoopPreludeCoda}{\LoopModel{p\dash b\x r\dash c}\xspace}

\title{How Depth Growth and Looping Shape In-Context Learning in Language Models}

\author{
    \makebox[\textwidth][c]{%
    \textbf{Ferdinand Kapl}\textsuperscript{\mdseries 1,2,3}\footnotemark[1] \quad
    \textbf{Kaitlin Maile}\textsuperscript{\mdseries 4}\footnotemark[2]\quad
    \textbf{Stefan Bauer}\textsuperscript{\mdseries 1,2,3}\footnotemark[2]\quad
    \textbf{Johannes von Oswald}\textsuperscript{\mdseries 4}\footnotemark[2]\quad
    }\\[0.4em]
    \makebox[\textwidth][c]{%
    \textsuperscript{1}\,Technical University of Munich \quad
    \textsuperscript{2}\,Helmholtz AI, Munich}\\
    \makebox[\textwidth][c]{%
    \textsuperscript{3}\,Munich Center for Machine Learning (MCML)  \quad
    \textsuperscript{4}\,Google, Paradigms of Intelligence Team}
}

\iclrfinalcopy 
\begin{document}

\maketitle
\begingroup
\renewcommand{\thefootnote}{\fnsymbol{footnote}}
\footnotetext[1]{Correspondence: \texttt{ferdinand.kapl@tum.de}.}
\footnotetext[2]{Provided equal feedback and guidance.}
\endgroup

\begin{abstract}
Looped and depth-grown language models reuse computations across depth and have attracted interest for their reasoning capabilities. Building on prior suggestive evidence of improved in-context-learning (ICL), we investigate how such reuse favors ICL, including how it shapes the acquisition, retention, and mechanisms of this ability. Our analysis spans pretrained language models up to 3B parameters and controlled tasks. The experiments provide evidence for a shared inductive bias toward learning from context: grown and looped models rely more on information in context and, compared with a full-depth dense reference, acquire ICL earlier in controlled tasks and retain it more strongly when learning fixed associations in the weights is also possible. Tracing attention heads through training reveals that depth growth propagates early induction heads across layers. Ablations show that grown and looped models rely more strongly on induction-ranked heads for both few-shot performance and later-token prediction. Together, these findings provide evidence for a shared inductive bias toward learning from context and suggest that (soft) parameter reuse shapes how this ability develops and which attention mechanisms support it.
\end{abstract}

\section{Introduction}

How a language model acquires its depth can shape the computations it learns. \emph{Depth growing} trains an initially shallow network and progressively duplicates blocks of layers, allowing later training stages to reuse computations learned earlier. \emph{Looping} instead repeatedly applies a shared block within each forward pass. Originally motivated by training efficiency \citep{fahlman1989cascade} and parameter efficiency, respectively, both approaches have attracted renewed interest for their reasoning capabilities \citep{saunshi2024inductive,saunshi2025reasoning,geiping2025scaling}. Although grown layers become independent after duplication, whereas looped layers remain tied, the two approaches can produce related patterns of iterative computation across depth \citep{kapl2026do,kapl2026from}.

This reuse of computation may also favor \emph{in-context learning} (ICL), as suggested by initial evidence in both grown and looped models \citep{kapl2026from}. ICL is the ability to use information supplied in a prompt or sequence without updating the model's parameters \citep{brown2020language}. A model that repeatedly refines its representations might also be better equipped to extract and apply patterns from context. Yet,  improved reasoning performance alone does not establish this ability. Predictions can rely on associations stored in the weights, on information inferred from the current sequence, or on both. Moreover, a model's improvement in predicting later tokens and its ability to learn a task from demonstrations are related but distinct measurements, with potentially different underlying mechanisms \citep{olsson2022context,yin2025which}.

Prior work provides promising evidence for a connection between depth reuse and ICL. Looped Transformers can learn synthetic regression and other data-fitting tasks with substantially fewer parameters than standard Transformers \citep{yang2024loopedtransformersbetterlearning,vonoswald2023transformers}. Studies of pretrained language models also report improved adaptation from demonstrations in grown and looped models for synthetic reasoning tasks \citep{kapl2026from}. Concurrently, \citet{wang2026smelt} report stronger gains with increasing context length and demonstration count in compute-matched looped mixture-of-experts models. What remains less understood is how these behavioral differences arise during training: whether growing and looping favor a context-based solution when memorization is also possible, how they affect the persistence of that solution, and how changes in attention-head mechanisms relate to the different measures of ICL.

In this work, we investigate these questions through complementary experiments on natural language and controlled tasks. We compare dense baselines with depth-grown and looped models on language-model configurations with dense reference sizes from 360M to 3B parameters. We measure position-dependent token loss and scaling with the number of demonstrations for few-shot learning. We then use RegBench \citep{akyurek2024icll} to study learning previously unseen formal languages, and Omniglot-based tasks \citep{chan2022bursty,singh2023transient} to separate learning from context from learning fixed associations in the weights. Finally, we track induction and function-vector heads through training and intervene on them in the final models. Our contributions are the following:

\begin{itemize}
    \item \textbf{Grown and looped language models benefit more from context.} In the main 3B language-model comparison, both achieve larger early-to-late token-loss reductions than the dense baseline, with the grown model also attaining lower absolute late-token loss. Both models also achieve higher GSM8K accuracy across all demonstration counts. On RegBench, grown and looped models predict unseen formal languages more accurately, with their advantage emerging within the first string (\cref{fig:language-icl,fig:regbench-icl}).
    \item \textbf{Growing and looping favor earlier acquisition and stronger retention of ICL.} In controlled Omniglot tasks where both context and fixed class–label associations can solve training queries, grown and looped models retain higher ICL accuracy and exhibit lower in-weights learning (IWL) accuracy than the full-depth dense reference. When labels are reassigned in every sequence, removing a fixed class–label solution, both acquire ICL earlier than the same reference (\cref{fig:omniglot-icl}).
    \item \textbf{Depth growth propagates early induction heads across layers.} Tracing final induction heads backward through training shows that strong heads in the grown model predominantly descend from heads already exhibiting strong induction behavior in early growing stages. Layer duplication produces repeated patterns of this behavior across depth, while looped models exhibit analogous patterns across successive applications of their shared looped block (\cref{fig:language-emergence,fig:head-trajectories-locations}).
    \item \textbf{Grown and looped models rely more strongly on induction heads for ICL.} In the main language-model comparison, ablating induction-ranked heads causes larger drops in few-shot performance and larger increases in late-token loss than in the dense baseline.
    These interventions establish a greater contribution from induction-ranked heads to both behaviors, while function-vector heads remain important (\cref{fig:head-ablation-combined}).
\end{itemize}

Together, these experiments show that constructing depth through (soft) parameter reuse shapes both the development of context use and the attention mechanisms supporting it.

\section{Related Work}

\paragraph{Grown Models.}
Model growth reduces training cost by starting with a smaller network \citep{fahlman1989cascade}. Language-model approaches use layer copying and progressive stacking \citep{gong2019efficient,reddi2023efficient,kim2023solar,du2024stacking}, learned transformations between model sizes \citep{wang2023learning}, or masked and function-preserving expansions \citep{yao2023masked,chen2015net2net,gesmundo2023composable}. Growth operators also shape model capabilities: \MD duplicates a central block of layers and improves reasoning at comparable language-model capabilities \citep{saunshi2024inductive}. \citet{kapl2026do} find stronger reliance on later layers in grown models and propose \LD, which duplicates layers around the network's center independently of existing block boundaries.

\paragraph{Looped Models.}
Cross-layer parameter sharing provides computational depth with fewer weights, as in Universal Transformers \citep{dehghani2019universaltransformers}, recurrent stacking for translation \citep{dabre2019recurrent}, and ALBERT \citep{lan2020albertlitebertselfsupervised}, while \citet{takase2023lessons} study its design choices. Extensions add layer-specific low-rank adapters \citep{bae2025relaxed} or mixture-of-experts capacity \citep{csordas2024moeut}. Recurrence has also been studied for algorithmic length generalization \citep{fan2025loopedtransformerslengthgeneralization}, latent reasoning \citep{saunshi2025reasoning}, and inference-time computation scaling \citep{geiping2025scaling,zhu2025scaling}. Other work retrofits recurrence into pretrained models \citep{mcleish2025teachingpretrainedlanguagemodels,koishekenov2025encodethinkdecodescaling} or explores hierarchical and compact recursive architectures for iterative refinement \citep{wang2025hierarchicalreasoningmodel,jolicoeurmartineau2025morerecursivereasoningtiny}.

\paragraph{In-Context Learning.}
ICL enables adaptation from demonstrations without parameter updates \citep{brown2020language}. Studies show that Transformers infer unseen functions from examples \citep{garg2022what}, learn gradient-descent-like computations in restricted regression settings \citep{vonoswald2023transformers,gatmiry2024can}, and can be programmed to execute learning algorithms in context \citep{giannou2023looped}. Data properties such as burstiness and dynamic meanings encourage ICL \citep{chan2022bursty}, which can decline as persistent associations are learned in the weights \citep{singh2023transient}. Other work connects regular-language learning to in-context $n$-gram heads on RegBench \citep{akyurek2024icll} and studies context-dependent representations and their competition with pretrained semantics through graph tracing \citep{park2025iclr}. These settings let us probe context use with and without competing in-weights solutions.

Induction heads support pattern copying and are associated with improved later-token prediction \citep{elhage2021mathematical,olsson2022context}. Their formation has been studied in controlled synthetic tasks \citep{singh2024what} and language-model pretraining \citep{aoyama2026predicting}. Function-vector heads transmit task representations from demonstrations \citep{todd2024function}. \citet{yin2025which} find that few-shot performance depends primarily on these heads, many of which first exhibit induction behavior.

Looped Transformers can match standard Transformers on synthetic data-fitting tasks with substantially fewer parameters \citep{yang2024loopedtransformersbetterlearning}. Concurrent work \citep{wang2026smelt} reports gains that increase with sequence length and demonstration count in looped mixture-of-experts language models under matched compute, parameter, and KV-cache budgets .

Most closely related, \citet{kapl2026from} identify shared depth-wise computational patterns in grown and looped models, show that grown models can benefit from inference-time repetition, and report improved adaptation from demonstrations on reasoning primitives. We extend this evidence to ICL acquisition, retention, and mechanisms across pretrained language models and controlled tasks. We test whether growing and looping favor context use when memorization is possible, trace induction heads through layer duplication, and ablate induction and function-vector heads to assess their contributions to few-shot learning and token prediction.

\section{In-Context Learning of Depth-Grown and Looped Models}
We first examine how growing and looping affect learning from context. After introducing the models and measures (\cref{subsec:models-measures}), we evaluate context use in language models (\cref{subsec:language-context-use}), adaptation to unseen formal languages on RegBench (\cref{subsec:regbench}), and ICL with and without an in-weights solution on Omniglot (\cref{subsec:omniglot}).

\begin{figure}
    \centering
    \includegraphics[width=\linewidth]{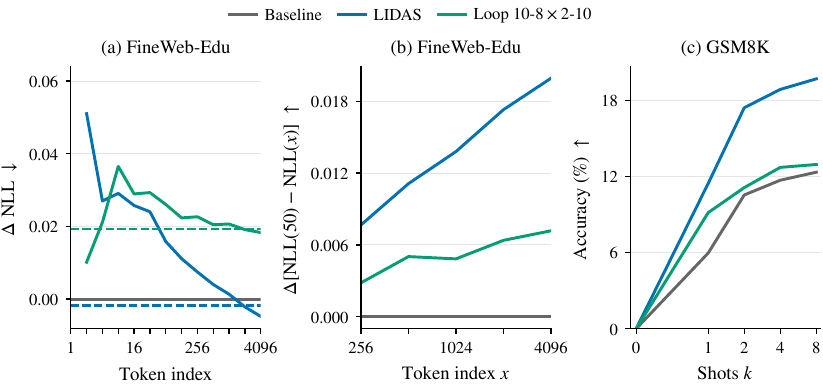}
    \caption{\textbf{Grown and looped models benefit more from context.} In the 3B comparison, grown and looped models show greater context gains in (a) NLL relative to the dense baseline and (b) early-to-late loss reduction relative to the baseline per token position, although the looped model retains higher NLL. Both panels aggregate token losses in the same power-of-two position bins: (a) compares the binned loss directly with the baseline, whereas (b) first computes each model's improvement from a fixed early token position (50) to the bin ending at $x$ and then subtracts the baseline's improvement. Dashed horizontal lines show mean $\Delta$NLL over all positions. (c) Both models exceed the baseline's few-shot GSM8K \citep{cobbe2021training} accuracy, with larger gains for \LD. Results use 10,000 FineWeb-Edu documents and three demonstration seeds, respectively. Additional model sizes and exact experimental setup in \cref{app:language-context}.}
    \label{fig:language-icl}
\end{figure}

\subsection{Models and Evaluation Measures}
\label{subsec:models-measures}

A dense model trains $L$ independently parameterized Transformer layers from scratch. A depth-grown model instead begins with a shallower model and periodically duplicates a block of layers, after which the copies train independently. The main paper focuses on \LD \citep{kapl2026do}, which selects the central layers without respecting earlier block boundaries. Comparisons with \MD \citep{saunshi2024inductive} and at other model sizes (360M and 1.7B) appear in \cref{app:cross-size-comparison}. The language models start with block size $b=4$ layers and add $b$ layers at each growth event, copying both weights and optimizer moments while retaining the training schedule.

A looped model reuses the same block of layers several times within each forward pass, with parameters remaining tied throughout training. We write \LoopPreludeCoda for $p$ prelude layers, a block of $b$ layers applied $r$ times, and $c$ coda layers, borrowing the naming conventions by \citet{geiping2025scaling}. Its effective depth is $p+br+c$, with $p+b+c$ unique layers. Our main 3B comparison uses a 36-layer dense reference \citep{bakouch2025smollm3}, a \LD{}-grown model reaching that depth, and the \LoopThree model, which has 36 layer applications and 28 unique layers. All three models are trained on the same 100B tokens in the same order. Within each size, all models share the data and optimizer-step budget and have matched final inference FLOPs. Growing uses shallower networks during part of training, saving an estimated 26\% of training FLOPs at 3B and 22\% at both 360M and 1.7B. Looping reduces unique parameters while retaining the same effective depth. \cref{app:lm-training} reports the configurations and training schedules.

\textbf{In-Context Learning Measures.}
We evaluate language models using next-token negative log-likelihood (NLL, lower is better), a token-loss ICL-Score \citep{olsson2022context}, and few-shot task performance. The ICL-Score measures the reduction in mean NLL from early positions 40--60 ("50") to later positions 490--510 ("500"), with higher values indicating greater early-to-late improvement. We use these token-loss windows to smooth potential single token position spikes and refer to the positions in the following by a single position for simplicity. In \cref{fig:language-icl}(b), we instead average later losses over power-of-two bins ending at $x$, keeping the early reference fixed. This measures improvement within each model, while we report individual token loss separately. Few-shot performance is measured through benchmark accuracy or token-overlap F1. Full protocols are given in \cref{app:lm-evaluation}.

\subsection{Context Use in Language Models}
\label{subsec:language-context-use}

We first compare how the different models use additional context in sequence modeling and few-shot tasks. These evaluations distinguish improvement with more context from absolute performance relative to the dense baseline. For sequence modeling, the depth-grown and looped models are worse at predicting early tokens compared to the baseline (\cref{fig:language-icl}(a)), with the biggest gap for very early tokens (token index $\le8$). After this peak, both models improve the prediction of tokens by lowering their NLL more, making better use of more tokens in the context. This is especially apparent when measuring the improvement of later tokens versus a fixed early position (50) and comparing it to the improvement of the baseline (\cref{fig:language-icl}(b)). The looped model, which uses significantly fewer unique parameters, retains higher NLL than the baseline even at the end of the sequence. \LD matches the baseline around position 1000 and achieves lower NLL thereafter (solid line) and on average across positions (dashed line).

On GSM8K \citep{cobbe2021training} few-shot accuracy in \cref{fig:language-icl}(c), the looped model, again using fewer parameters, keeps its advantage over the baseline across demonstration counts, but the baseline almost closes the gap at eight shots. The depth-grown model instead widens its performance improvement gap with additional examples, ending up around $+7\%$ above the baseline. GSM8K evaluates final answers extracted from generated solutions to new questions, so these gains reflect improved answer generation from in-context demonstrations, a setting relevant to practical few-shot prompting. As an additional comparison, Appendix \cref{fig:appendix-fewshot-subsets} reports separate averages over five multiple-choice and three generation tasks. At 3B, \LD improves on the baseline in both subsets, while the looped model's advantage is concentrated in generation.

\subsection{In-Context Learning Languages: RegBench Dataset}
\label{subsec:regbench}

RegBench \citep{akyurek2024icll} tests adaptation to an unseen formal language. Each input concatenates strings sampled from one deterministic finite automaton (DFA), with held-out automata used for testing. We measure total variation distance (TVD) between the predicted next-symbol distribution and the uniform distribution over legal continuations. All models apply twelve layers per forward pass, with two unique layers for the looped model.

\begin{figure}
    \centering
    \includegraphics[width=\linewidth]{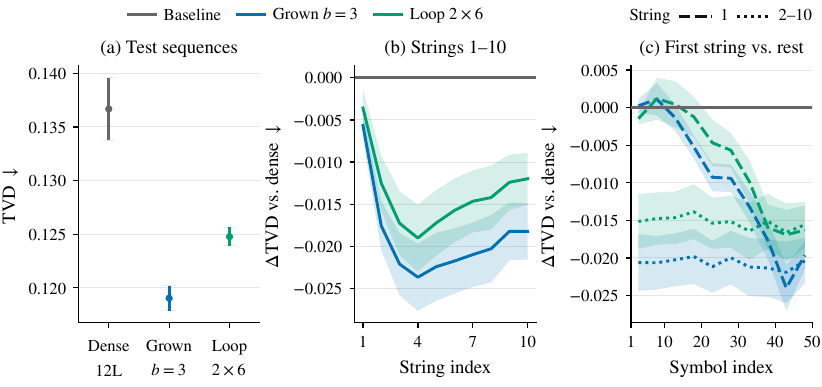}
    \caption{\textbf{Grown and looped models predict unseen formal languages more accurately.} In-context language learning on RegBench \citep{akyurek2024icll}. (a) Test total variation distance (TVD) for the dense 12-layer, \LoopTwoSix, and Grown $b=3$ model (mean $\pm$ standard error over four training seeds). (b) TVD relative to the dense model against in-context string index. (c) The same comparison by symbol index, separately for the first string and strings 2--10. Lower is better for all panels. Bands show standard errors. The looped model has width 256, versus 128 for dense and grown models: hyperparameters are selected on validation data, rather than matched in width. Panel (a) uses the full test set. Panels (b,c) retain only string indices present in every test sequence (strings 1--10). Experimental setup and full details in \cref{app:regbench}.}
    \label{fig:regbench-icl}
\end{figure}

In \cref{fig:regbench-icl}(a), the grown and looped models achieve lower TVD than the dense baseline on novel languages. Their advantage increases over the first few strings, peaking around the fourth (\cref{fig:regbench-icl}(b)). When zooming in on the test sequences in \cref{fig:regbench-icl}(c), both alternative training variants adapt faster to the new language by predicting symbols (akin to characters or tokens) within the first string better after roughly 15 symbols. Additionally, on average, after the first string the grown and looped model do not continue to increase their advantage with more symbols but keep a constant improvement. Appendix \cref{fig:appendix-regbench} extends the comparison including DFA accuracy and additional grown models also evaluated with four training seeds, with the precise setup in \cref{app:regbench}.

\begin{figure}
    \centering
    \includegraphics[width=\linewidth]{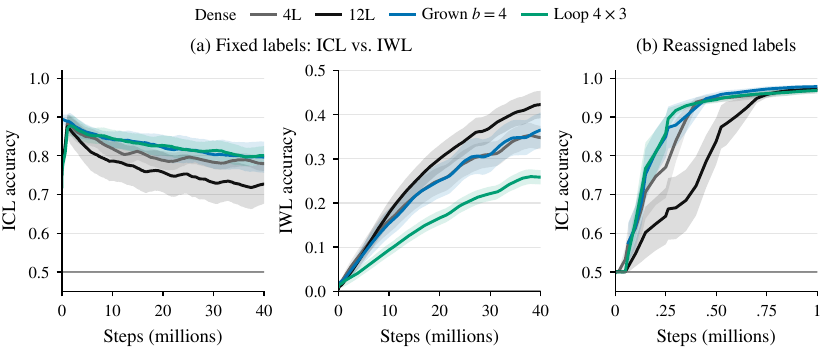}
    \caption{\textbf{Growing and looping favor stronger retention of ICL, at the cost of in-weights learning (IWL), and earlier acquisition of ICL when IWL is not possible.} Solution preference and emergence of in-context learning on Omniglot-based tasks \citep{singh2023transient}. (a) Training uses fixed class labels and $P(\mathrm{bursty})=1$, so both context and in-weights associations can solve each training query. ICL is measured on held-out classes with labels reassigned per sequence. IWL is measured on training classes whose query class is absent from context. (b) Training follows the ICL probe setup using images from training classes, with labels reassigned in every sequence to remove a fixed class--label solution. Curves show mean accuracy, with bands indicating standard error across four (a) and eight (b) smoothed seed trajectories. Growth occurs at 50k and 100k steps. Experimental details in \cref{app:omniglot}.
    }
    \label{fig:omniglot-icl}
\end{figure}

\subsection{In-Context and In-Weights Learning: Omniglot}
\label{subsec:omniglot}

We next examine learning from context when fixed associations can also be stored in the weights, or when labels vary between sequences, removing the in-weights learning (IWL) solution. For that purpose, we use the Omniglot image-classification setup of \citet{chan2022bursty}. Each sequence contains eight image--label pairs followed by a query image. In the fixed-label setting, every training class has a persistent label and the query class always appears among the context examples, with $P(\mathrm{bursty})=1$, meaning that every training sequence contains multiple demonstrations of the same classes. The training query can therefore be answered either by matching it to the current examples in context or by learning its class--label association in the weights. We distinguish these capabilities with two probes: ICL accuracy on held-out classes with labels reassigned in each sequence, and IWL accuracy on training classes whose query class is absent from context. We compare dense models with 4 and 12 layers, a \LoopFourThree model, and a \LD{}-grown model that expands from 4 to 8 layers at 50k steps and to 12 layers at 100k steps. All models have the same hidden width and use the same optimizer and batch size. We average over four training seeds for the fixed-label setting and eight for the ICL-only setting below. \cref{app:omniglot} gives the full setup.

\cref{fig:omniglot-icl}(a) shows an early rise in ICL followed by a gradual decline as IWL improves, reflecting the transience of ICL observed by \citet{singh2023transient}. Over the shared training horizon, the grown and looped models retain higher mean ICL accuracy but lower IWL accuracy than the dense 12-layer model. The grown model has similar ICL and IWL accuracies to its four-layer reference, from which it first grows at 50k steps. The looped model attains similar ICL accuracy to the grown model but significantly lower IWL accuracy. These trajectories are consistent with growing and looping favoring learning from context relative to the full-depth dense model at the same effective depth. For looped models, this can come at the cost of weaker in-weights learning, while the grown model appears to inherit both capabilities from its shallow starting model. 

To test acquisition when a fixed class--label solution is unavailable (ICL-only), we train a different set of models with labels 0 and 1 reassigned independently in every sequence, mirroring the ICL probing setup on training images. Each sequence contains two classes with four examples each, followed by a query from one of them. In \cref{fig:omniglot-icl}(b), ICL accuracy emerges earlier for the looped, grown, and four-layer dense models than for the dense 12-layer model. By 1M steps, all four tested configurations almost reach 100\%. Thus, this setting does not differentiate how much ICL is attained but rather the timing and consistency of ICL emergence, given that ICL is the only correct solution to the training problem.

\section{Growing and Looping Increase Reliance on Induction Heads}

The behavioral results suggest that growing and looping favor learning from context. Returning to language models, we examine the attention mechanisms supporting this behavior. We first track induction behavior during training and across layers (\cref{subsec:induction-development}), then ablate heads to test their contribution to context use (\cref{subsec:head-ablation}).

\subsection{Depth Growth Propagates Early Induction Heads}
\label{subsec:induction-development}

\begin{wrapfigure}[20]{r}{1.74in}
    \centering
    \includegraphics[width=\linewidth]{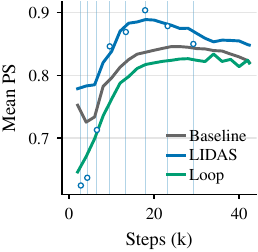}
    \caption{\textbf{Induction behavior during training.} Top-2\% mean PS (3B), with heads reselected at each checkpoint. Vertical lines mark growth. Hollow circles show immediate post-growth scores.}
    \label{fig:language-emergence}
\end{wrapfigure}

\textbf{Attention-Head Measures.}
We characterize attention heads using prefix-matching (PS) and function-vector (FV) scores. PS measures attention to the token following an earlier occurrence in repeated random-token sequences \citep{olsson2022context}. FV scores measure how much replacing a head's output with its clean task mean restores correct-answer prediction when demonstration labels are shuffled \citep{todd2024function}. We use these scores to select heads for ablation. Full evaluation protocols are given in \cref{app:induction-measures}. We first compare induction behavior during language-model pretraining by averaging PS over the top 2\% of heads at each checkpoint, reselecting them as training progresses. Checkpoint and aggregation details are given in \cref{app:lm-checkpoints,app:induction-measures}.

In \cref{fig:language-emergence}, \LD develops higher PS than the dense baseline and finishes training with a higher score, while the looped model finishes close to the baseline. Early growth events temporarily disrupt prefix matching, followed by quick recovery during further training. These aggregate scores motivate tracing the heads responsible for the grown model's strong induction behavior. Full trajectories of PS, function-vector scores, few-shot performance, and the token-loss ICL-Score appear in Appendix \cref{fig:appendix-trajectories}.

\begin{figure}[!t]
    \centering
    \includegraphics[width=\linewidth]{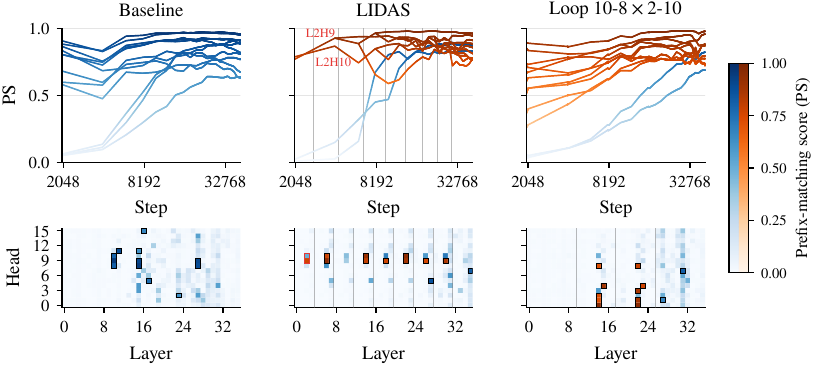}
    \caption{\textbf{Depth growth propagates early strong induction heads across layers.} Development and final location of individual induction heads. \textbf{Top:} smoothed prefix-matching scores (PS) of each 3B model's twelve final top-2\% heads over training (logarithmic step axis). \LD lineages follow layer duplication, with shared ancestral paths drawn once. In \LD, the twelve final heads trace back to four ancestors in the initial four-layer model. Two of these, L2H9 and L2H10, already have high PS before growth and account for ten of the twelve final heads. \textbf{Bottom:} final PS for every layer and head (zero-based indices). Among the final top-2\% PS heads, orange highlights descendants of the early L2H9/L2H10 heads in \LD and heads within the repeated block in \LoopThree. Other heads are blue. Black outlines identify final top-2\% heads, except the original L2H9 in \LD, which shares a red outline with L2H10. Vertical lines mark growth events (top), four-layer intervals in \LD and loop-block boundaries (bottom). Selection, plotting, and evaluation details in \cref{app:head-lineages}. FV heads show analogous repetition but emerge later (Appendix \cref{fig:appendix-fv-trajectories-locations}).}
    \label{fig:head-trajectories-locations}
\end{figure}

We next follow the twelve final top-2\% PS heads in each 3B model through training (\cref{fig:head-trajectories-locations}, top). These identities remain fixed, unlike the checkpoint-wise selection for the average in \cref{fig:language-emergence}. For \LD, we trace the heads backward through layer duplication to the starting four-layer model. Copies share an ancestral trajectory until duplication, so we draw that shared path once.

The twelve final \LD heads trace back to just four ancestors. Two of these, L2H9 and L2H10 (zero-based indices), already have high PS before the first growth event and account for ten of the twelve final heads, highlighted in orange. The remaining two originate from initially low-scoring heads whose induction behavior develops later. Thus, most of the strongest final induction heads arise from copying and further training early induction heads. The original L2H10 retains high PS but is outside the final top-2\% heads, while three of its copies are included.

The final locations in \cref{fig:head-trajectories-locations} (bottom) connect these lineages to repeated patterns across depth. In \LD, strong heads recur at the same head index in layers separated by multiples of the four-layer growth block, extending from early layers ($L<8$) into later layers. In \LoopThree, high-PS heads likewise recur across the two applications of the eight-layer shared block, highlighted in orange. These are separate applications of tied heads, whereas the grown model's copies train independently. The dense baseline shows no comparable pattern of repetition by block size.

FV heads likewise recur at the grown or looped block size, but emerge later than induction heads (Appendix \cref{fig:appendix-fv-trajectories-locations}). In \LD, two ancestors give rise to six of the twelve final top-2\% FV heads.

\subsection{Induction Heads Contribute More to Context Use}
\label{subsec:head-ablation}

The trajectories and locations reveal how induction behavior develops and repeats across depth. A high prefix-matching score alone does not establish a head's importance for ICL. We therefore ablate heads in the final models to test their contributions to few-shot performance and token loss.

Following the mean-ablation protocol of \citet{yin2025which}, we rank final-checkpoint heads by PS or FV score and replace the outputs of the selected 1--20\% with mean activations. For looped models, each head application is counted and ablated separately. Here we focus on exclusive induction rankings, which first exclude the top 2\% of FV heads, and exclusive FV rankings, which first exclude the top 2\% of induction heads. We evaluate changes from each model's unablated few-shot ICL (F1) and ICL-Score. \cref{app:head-ablation} specifies the replacement means, indexing, and additional random controls.

\begin{figure}[!t]
    \centering
    \includegraphics[width=\linewidth]{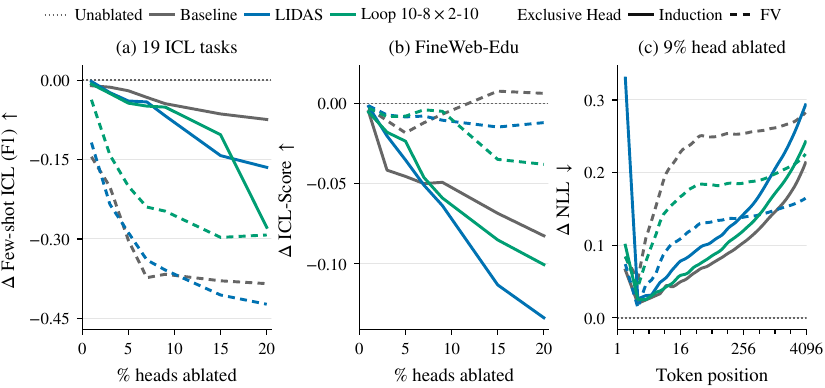}
    \caption{\textbf{Grown and looped models rely more on induction heads than the baseline for few-shot ICL and later-token prediction.} Mean ablation in the 3B baseline, \LD, and \LoopThree. Solid lines ablate heads ranked by PS after excluding the top-2\% FV heads. Similarly, dashed lines show FV-head ablations. (a) Change in few-shot ICL (F1) on 19 tasks. (b) Change in ICL-Score on FineWeb-Edu. Both changes are ablated minus the same model's unablated score (lower is worse). At larger ablation fractions, induction-head ablation reduces both scores more in grown and looped models than in the baseline, although FV-head ablation hurts few-shot ICL more in all three models. (c) Position-wise NLL increase from each model's unablated reference at 9\% ablation over the 4096-token context (higher is worse). Curves are log-binned and smoothed. FV-ablation effects largely plateau, while induction-ablation effects grow with position and eventually exceed FV-ablation effects in \LD and \LoopThree. Additional ablation fractions appear in Appendix \cref{fig:appendix-ablation-position} with experimental details in \cref{app:head-ablation}.}
    \label{fig:head-ablation-combined}
\end{figure}

Consistent with observations by \citet{yin2025which}, all three models in \cref{fig:head-ablation-combined}(a,b) show greater few-shot ICL degradation from FV-head ablation and greater ICL-Score (token-loss difference) reduction from induction-head ablation. At larger ablation fractions, exclusive induction-head ablation reduces few-shot performance and ICL-Score more in grown and looped models than in the baseline, indicating greater reliance on these heads. FV heads remain important in all models, but their ablation effects relative to the baseline vary across models and sizes for this comparison (Appendix \cref{fig:appendix-exclusive-ablation}). 

\cref{fig:head-ablation-combined}(c) shows the NLL increase by token position at 9\% ablation. Beyond the first few tokens, FV-head ablation hurts early-token prediction more than induction-head ablation. The FV-ablation effect then largely plateaus, while the induction-ablation effect grows with position. Compared with the baseline, \LD and \LoopThree show smaller FV-ablation effects across most positions and larger induction-ablation effects at later positions. Within both models, induction heads eventually become more important than FV heads for later-token prediction, whereas FV heads remain more important in the baseline. Together, these results indicate that both head types support context use, with greater reliance on induction heads in grown and looped models.

\section{Conclusion}
Our experiments provide evidence that depth growing and looping favor learning from context. In the main language-model comparison, both achieve larger early-to-late token-loss reductions and higher GSM8K accuracy than the dense baseline. Grown and looped models also predict unseen formal languages more accurately. In controlled classification tasks, both acquire ICL earlier than a full-depth dense reference when labels are reassigned between sequences, and retain higher ICL accuracy when fixed associations can also be learned in the weights. Mechanistically, depth growth propagates early induction heads through layer duplication, with most of the strongest final heads descending from a small set of early ancestors. This produces repeated patterns across depth analogous to those in looped models, even though grown layers train independently after copying. Ablations in the main language-model comparison show that grown and looped models rely more on induction-ranked heads for few-shot performance and later-token prediction. Function-vector heads remain more important for few-shot performance in all models, suggesting that depth reuse shifts the relative contributions of these mechanisms. Together, these findings suggest that reusing computations across depth shapes how models acquire and retain ICL, as well as the attention mechanisms supporting it.

The controlled tasks suggest that depth reuse shifts the competition between in-context and in-weights learning: grown and looped models retain higher ICL accuracy while attaining lower IWL accuracy than the full-depth dense reference. One possible explanation is that copying early context-dependent computations through growth, or repeatedly applying them through shared layers, makes a context-based solution less susceptible to being displaced as fixed associations are learned in the weights.

\FloatBarrier

\subsection*{Acknowledgments}
The authors would like to thank Nino Scherrer for insightful discussions and support throughout this work. This work was partially supported by the Helmholtz Foundation Model Initiative and the
Helmholtz Association. The authors thank the Gauss Centre for Supercomputing e.V. (GCS) for providing computing time on the Supercomputer JUPITER at the Jülich Supercomputing Centre via the John von Neumann Institute for Computing. JUPITER is supported by the EuroHPC JU and GCS through funding by the European Commission, the German Federal Ministry of Research, Technology and Space, and the Ministry of Culture and Science of the State of North Rhine-Westphalia. Furthermore, the authors appreciate the computational resources provided by the National High Performance Computing Centre (www.nhr.kit.edu). The
research presented is supported by the TUM Georg Nemetschek Institute Artificial Intelligence for
the Built World.

\subsection*{AI use statement}

Generative AI tools, including LLM-based coding agents, supported the implementation of our methods and experiments. We further used them to aid the writing of plotting code behind our figures and to polish the wording of the manuscript. All AI-assisted code, figures, and text were checked by the authors, and the research questions, experimental design, and conclusions are our own. We take full responsibility for the content of this work, including everything produced with the help of generative AI.

\subsection*{Reproducibility statement}

The information needed to reproduce our experiments is provided in the main text and appendix. Language-model architectures, training configurations, and growth schedules are described in \cref{app:lm-training}. Evaluation protocols and mechanistic analyses are detailed in \cref{app:language-context,app:induction,app:head-ablation}, and the RegBench and Omniglot setups, hyperparameters, and data generation procedures in \cref{app:regbench,app:omniglot}. We use publicly available datasets and build on public implementations, including the nanotron training framework (\url{https://github.com/huggingface/nanotron}).

\bibliography{iclr2027_conference}
\bibliographystyle{iclr2027_conference}

\newpage

\appendix
\section{Language Model Training and Configurations}
\label{app:lm-training}

\subsection{Training Setup}
\label{app:lm-pretraining}

We train autoregressive, decoder-only Transformers with a next-token cross-entropy objective. The 360M and 1.7B configurations follow SmolLM \citep{benallal2024smollmcorpus}, and the 3B configuration follows SmolLM3 \citep{bakouch2025smollm3}. All use pre-RMSNorm, SwiGLU feed-forward layers, tied input and output embeddings, and no biases, dropout, or query/key normalization. \cref{tab:lm-configs} summarizes the dense reference configurations. Model sizes refer to these references; looped models have fewer unique parameters, and grown models reach the reference depth only during training. Comparisons between methods share the data and optimization settings within each size. The sizes differ in training budget, and the 3B configuration also changes the data, tokenizer, and context length, so they test robustness of observations across all these differences.

\begin{table}[htbp]
    \centering
    \small
    \caption{Language-model training configurations.}
    \begin{tabular}{lrrr}
        \hline
        & \textbf{360M} & \textbf{1.7B} & \textbf{3B} \\
        \hline
        Dense layers $L$ & 32 & 24 & 36 \\
        Hidden / MLP width & 960 / 2560 & 2048 / 8192 & 2048 / 11008 \\
        Query / key-value heads & 15 / 5 & 32 / 32 & 16 / 4 \\
        Vocabulary size & 49,152 & 49,152 & 128,256 \\
        Context length $C$ & 2048 & 2048 & 4096 \\
        Rotary base & $10^4$ & $10^4$ & $5\times10^4$ \\
        RMSNorm $\epsilon$ & $10^{-5}$ & $10^{-5}$ & $10^{-6}$ \\
        Global batch (sequences) & 512 & 1024 & 576 \\
        Training steps & 200,000 & 200,000 & 42,400 \\
        Training tokens (billions) & 209.7 & 419.4 & 100.0 \\
        Peak learning rate & $3\times10^{-3}$ & $5\times10^{-4}$ & $2\times10^{-4}$ \\
        Linear warm-up (steps) & & 2000 & \\
        AdamW $(\beta_1,\beta_2)$ & & $(0.9,0.95)$ & \\
        AdamW $\epsilon$ & & $10^{-8}$ & \\
        Gradient clipping norm & & 1 & \\
        Weight decay & 0.01 & 0.01 & 0.1 \\
        Embedding weight decay & 0.01 & 0.01 & 0 \\
        Weights / activations & & bf16 & \\
        Gradient accumulation & & fp32 & \\
        Base initialization std.\ $\sigma$ & 0.0417 & 0.0221 & 0.02 \\
        \hline
    \end{tabular}
    \label{tab:lm-configs}
\end{table}

\paragraph{Data and positional encoding.}
The 360M and 1.7B models use the SmolLM-corpus mixture of FineWeb-Edu-dedup, Cosmopedia-v2, Python-Edu, and Open-Web-Math \citep{benallal2024smollmcorpus}, contributing approximately 70\%, 15\%, 9\%, and 6\% of training tokens, respectively. They use the cosmo2 tokenizer and rotary position embeddings in all layers. The 3B models use the SmolLM3 tokenizer and its 100B-token ablation recipe. The initial mixture includes FineWeb-Edu, DCLM, multilingual FineWeb-2, scientific text, code, and mathematics. At step 36,040, training switches to the decay mixture, with increased emphasis on mathematics, code, and synthetic reasoning. All models use shuffled, packed sequences of the lengths in \cref{tab:lm-configs}; 3B additionally masks attention across document boundaries and omits rotary embeddings in every fourth layer.

\paragraph{Optimization and initialization.}
All models use AdamW with the hyperparameters in \cref{tab:lm-configs}. After warm-up, learning rates stay constant, then decay to zero using $1-\sqrt{\cdot}$ decay from step 170,000 (360M/1.7B) or linear decay from step 36,040 (3B). Matrix weights use zero-mean normal initialization with base standard deviation $\sigma$; RMSNorm scales start at one. Attention output and MLP down projections additionally receive a $1/\sqrt{2L_{\mathrm{init}}}$ factor, where $L_{\mathrm{init}}=4$ for grown models and equals the effective depth for dense and looped models. The 360M looped model omits this factor, so its comparisons also include an initialization difference.

\subsection{Growing and Looping}
\label{app:lm-architectures}

\paragraph{Dense and grown models.}
The dense baseline has $L$ independently parameterized layers throughout training. Grown models start with $b=4$ layers and repeatedly duplicate four contiguous layers until reaching $L$. A duplication copies both the layer weights and Adam's first and second moments. The copies become independent trainable layers immediately; embeddings are retained, and neither the global step nor the learning-rate schedule is reset. \MD \citep{saunshi2024inductive} selects the middle four-layer block, choosing the block just before the center when the number of blocks is even. \LD \citep{kapl2026do} instead selects the middle four layers without respecting previous block boundaries. For example, \LD maps an eight-layer stack to
\[
 [0,1,\:\mathbf{2,3,4,5},\:6,7]\ \longmapsto\ [0,1,\:\mathbf{2,3,4,5,2,3,4,5},\:6,7].
\]
Layer indices in this expression identify the source of the copied parameters, not persistent weight sharing.

We use the Prop-1 schedule \citep{saunshi2024inductive}, allocating steps in proportion to depth. For $K=L/b$ stages, stage $i$ has depth $ib$ and receives
\[
T_i = \frac{i}{\sum_{j=1}^{K}j}\,T_{\mathrm{prop}}, \qquad i=1,\ldots,K,
\]
steps, up to integer rounding. The budget $T_{\mathrm{prop}}$ excludes learning-rate decay, which is appended to the final full-depth stage. For 360M/1.7B, $T_{\mathrm{prop}}=170{,}000$, including warm-up in the first stage. For 3B, the 2000 warm-up steps are also excluded from the proportional split and prepended to the first stage, giving $T_{\mathrm{prop}}=36{,}040-2000=34{,}040$. Only \LD is evaluated at 3B. At 1.7B, \MD branches from \LD{}'s eight-layer checkpoint at step 24,285: the two methods share the preceding trajectory, including the first growth event.

\paragraph{Looped models.}
We write \LoopPreludeCoda for $p$ prelude layers, a block of $b$ layers applied $r$ times, and $c$ coda layers. The effective depth is $p+br+c$, but only $p+b+c$ layers have unique parameters. Gradients from each application accumulate into the shared parameters. Prelude and coda are omitted from the notation when empty. The 1.7B comparison uses \LoopSeventeen: 24 layer applications, 16 unique layers, and approximately 1.17B parameters. The main 3B comparison uses \LoopThree: 36 applications, 28 unique layers, and approximately 2.45B parameters. Its eight-layer loop preserves the rotary/no-position-embedding role of each shared layer. The 360M comparison uses \LoopThreeSixty (32 applications, 20 unique layers, approximately 244M parameters). These comparisons match effective final depth and training tokens within a size; growing uses fewer layer applications during early training, and looping reduces the number of unique parameters.

\subsection{Checkpoints and Comparison Scope}
\label{app:lm-checkpoints}
\label{app:cross-size-comparison}

Final-checkpoint evaluations use step 200,000 for 360M/1.7B and step 42,400 for 3B. The 1.7B and 3B trajectories generally use checkpoints every 2000 steps, with additional saves at growth boundaries and at the final step. The 360M dense and grown trajectories use a 10,000-step grid, with additional growth saves. At a growth boundary, we distinguish the trained pre-growth checkpoint from the immediate post-growth checkpoint, before any further optimizer update. The latter is shown as a hollow marker at the same training step. Vertical rules mark the growth events.

The appendix compares dense, \MD, \LD, and the selected looped model at 360M and 1.7B, and dense, \LD, and the selected looped model at 3B. There is no 3B \MD run. All comparisons use one language-model training seed per arm; evaluation bands do not estimate variation across training runs. The configuration differences and 360M loop initialization exception in \cref{app:lm-pretraining} apply throughout.

\paragraph{Loop configurations in appendix figures.}
\label{app:loop-configurations}
A single green legend entry, ``Loop'', denotes \LoopThreeSixty at 360M, \LoopSeventeen at 1.7B, and \LoopThree at 3B. The size of the panel determines the configuration. The architecture and parameter-sharing convention are given in \cref{app:lm-architectures}.

\FloatBarrier

\section{Context Use in Language Models}
\label{app:lm-evaluation}
\label{app:language-context}

This section gives the evaluation protocols and results across sizes for general benchmarks, token loss, downstream few-shot performance, and graph tracing. Shared training configurations and the size-specific Loop definitions are given in \cref{app:lm-training}.

\subsection{General Benchmark Evaluations}
\label{app:general-benchmarks}

\cref{tab:general-benchmarks} reports general benchmark performance using OLMES \citep{gu2025olmes} at the final checkpoints: step 200,000 for 360M/1.7B and step 42,400 for 3B. We evaluate the nine-task core suite and MMLU \citep{hendrycks2021mmlu} with five fixed demonstrations and a maximum context length of 2048 tokens for every model, including the 3B models.

\begin{table}[htbp]
    \centering
    \caption{\textbf{General benchmark performance across model sizes.} Accuracy (\%), with higher values better. \textbf{9-task MC} is the unweighted mean over ARC-Easy, ARC-Challenge, BoolQ, CommonsenseQA, HellaSwag, OpenBookQA, PIQA, Social IQa, and WinoGrande; \textbf{MMLU} is the unweighted mean over its 57 subjects. Both columns use five-shot ranked-classification (RC) scoring of answer continuations with OLMES task-specific normalization and a 2048-token context limit. Here MC denotes multiple-choice benchmarks; scores use the RC formulation throughout, rather than answer-letter scoring. Model sizes denote dense reference configurations; loop layouts follow \cref{app:lm-architectures}. Best values within each size are bold.}
    \label{tab:general-benchmarks}
    \small
    \setlength{\tabcolsep}{10pt}
    \renewcommand{\arraystretch}{1.08}
    \begin{tabular}{@{}clrr@{}}
        \toprule
        Model & Strategy & 9-task MC $\uparrow$ & MMLU $\uparrow$ \\
        \midrule
        \multirow{4}{*}{\rotatebox[origin=c]{90}{360M}}
            & Baseline & 54.69 & \textbf{34.67} \\
        \cmidrule(lr){2-4}
            & \MD & 56.00 & 34.11 \\
            & \LD & \textbf{57.10} & 34.09 \\
            & \LoopThreeSixty & 54.21 & 33.69 \\
        \midrule
        \multirow{4}{*}{\rotatebox[origin=c]{90}{1.7B}}
            & Baseline & 62.97 & 40.12 \\
        \cmidrule(lr){2-4}
            & \MD & 64.61 & 40.43 \\
            & \LD & \textbf{64.69} & \textbf{40.86} \\
            & \LoopSeventeen & 63.25 & 39.89 \\
        \midrule
        \multirow{3}{*}{\rotatebox[origin=c]{90}{3B}}
            & Baseline & 61.33 & 39.15 \\
        \cmidrule(lr){2-4}
            & \LD & \textbf{62.89} & \textbf{39.28} \\
            & \LoopThree & 60.53 & 38.32 \\
        \bottomrule
    \end{tabular}
\end{table}

The core suite uses test splits for ARC-Easy, ARC-Challenge, and OpenBookQA, and validation splits for the remaining tasks. ARC-Easy, BoolQ, HellaSwag, PIQA, and Social IQa use the standard fixed 1000-example subsamples; ARC-Challenge, CommonsenseQA, OpenBookQA, and WinoGrande use their full splits of 1172, 1221, 500, and 1267 examples, respectively. MMLU uses all 14,042 test examples across its 57 subjects. Following OLMES, answer log-likelihoods are normalized by character count for ARC-Easy, HellaSwag, PIQA, Social IQa, and MMLU; corrected by unconditional answer log-likelihood for ARC-Challenge, CommonsenseQA, and OpenBookQA; and left unnormalized for BoolQ and WinoGrande.

\FloatBarrier

\subsection{Token Loss and the ICL Score}
\label{app:token-loss}

\begin{figure}[!htbp]
    \centering
    \includegraphics[width=\linewidth]{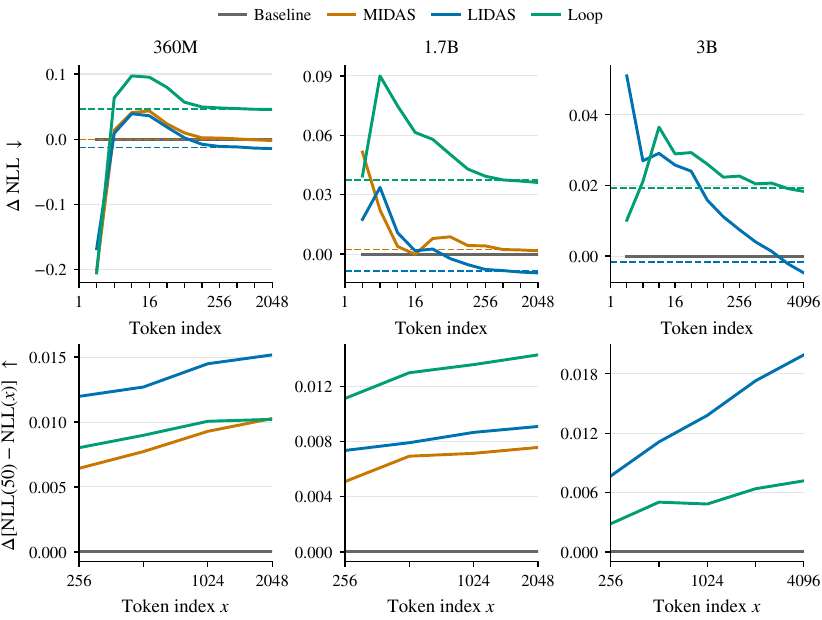}
    \caption{\textbf{Token loss and context gains across model sizes.} Columns show 360M, 1.7B, and 3B, using the position bins of \cref{fig:language-icl}. Top: final-checkpoint NLL relative to the same-size dense model in power-of-two position bins plotted at their upper endpoints, with matching-color dashed full-context means computed over raw positions. Bottom: the early-to-late loss reduction relative to dense (\cref{eq:icl-score-position}) for later bins ending at $x\in\{256,\ldots,C\}$. All curves use 10,000 long FineWeb-Edu documents. Native contexts are $C=2048/2048/4096$ positions; the evaluation protocol is given in \cref{app:token-loss}. Loop configurations are specified in \cref{app:loop-configurations}.}
    \label{fig:appendix-context}
\end{figure}

\cref{fig:language-icl}(a,b), \cref{fig:appendix-context}, and \cref{fig:appendix-trajectories} (bottom row) use a fixed set of 10,000 long FineWeb-Edu-dedup documents. After tokenization, we retain documents with at least $C+1$ tokens and truncate to $C+1$, yielding $C$ next-token predictions per document. Each forward pass therefore uses a single document and every position has the same document count. The 2048- and 4096-position evaluations use their model's native tokenizer and separately selected document sets.

For tokens $x_0,\ldots,x_C$, define
\begin{equation}
    \ell_j=-\log p(x_j\mid x_0,\ldots,x_{j-1}),\qquad j=1,\ldots,C,
    \label{eq:position-nll}
\end{equation}
and let $\bar\ell_j$ be its mean over documents, in nats. Position labels in the trajectory and position plots use this one-based prediction index. Evaluation uses bf16 forward passes. For the position-wise $\Delta$NLL panels in \cref{fig:language-icl}(a) and \cref{fig:appendix-context}, position means are grouped into non-overlapping power-of-two bins $(2^{y},2^{y+1}]$ for $y=0,\ldots,\log_2 C-1$, with position 1 added to the first bin. The integer position ranges are therefore 1--2, 3--4, 5--8, \ldots, $C/2+1$--$C$, giving 11 bins at $C=2048$ and 12 at $C=4096$; every position belongs to exactly one bin. Each bin mean is plotted at the bin's upper endpoint on a base-2 logarithmic axis. The plotted quantity is the model's binned NLL minus the same-size dense baseline's binned NLL. Thus, a negative value denotes better token prediction at those positions, rather than a larger improvement relative to the model's own early-token loss.

The dashed references in \cref{fig:language-icl}(a) and \cref{fig:appendix-context} average the raw position means with equal weight, before any position binning. With $D=10{,}000$ documents and $C$ scored prediction positions per document ($C=2048$ at 360M/1.7B and $4096$ at 3B), the reference for model $m$ is
\begin{equation}
    \begin{aligned}
    \overline{\Delta\mathrm{NLL}}_m
    &= \frac{1}{C}\sum_{j=1}^{C}\left(\bar\ell_j^{(m)}-\bar\ell_j^{(\mathrm{base})}\right) \\
    &= \frac{1}{DC}\sum_{d=1}^{D}\sum_{j=1}^{C}\left(\ell_{d,j}^{(m)}-\ell_{d,j}^{(\mathrm{base})}\right) \\
    &= \overline{\mathrm{NLL}}_m-\overline{\mathrm{NLL}}_{\mathrm{base}}.
    \end{aligned}
\end{equation}
This is the difference in token-averaged dataset NLL on this fixed long-document subset, using the same truncation and per-document context resets as the position curves. It is not an unweighted average of the logarithmic bin means, whose position counts differ.

Following the token-loss diagnostic of \citet{olsson2022context}, we summarize context use at each checkpoint by
\begin{equation}
    S_{\mathrm{ICL}}=\frac{1}{21}\sum_{j=40}^{60}\bar\ell_j
                   -\frac{1}{21}\sum_{j=490}^{510}\bar\ell_j.
    \label{eq:icl-score}
\end{equation}
A positive score indicates lower NLL at the later window. A change in this difference can arise from either window; it need not imply improved absolute prediction at late positions. Window-decomposition plots therefore report the two means separately. When subtracting the dense model's corresponding window loss along training, we use only steps at which both models have a saved checkpoint, without interpolating a baseline at growth steps. The combined head-ablation figure uses the same window definition and reports its change from the unablated model in \cref{fig:head-ablation-combined}(b), while \cref{fig:appendix-ablation} reports absolute scores.

\paragraph{Loss reduction across context positions.}
For the final-checkpoint comparison in \cref{fig:language-icl}(b) and \cref{fig:appendix-context}, we keep the early window at positions 40--60 and let the later window be the power-of-two bin ending at the target position $x$, $W_x=\{x/2+1,\ldots,x\}$, for $x\in\{256,512,1024,2048\}$ and additionally $4096$ at 3B. These are the same bins as in the $\Delta$NLL panels. We plot
\begin{equation}
    \Delta S_m(x)=\left[\frac{1}{21}\sum_{j=40}^{60}\bar\ell_j^{(m)}-\frac{1}{|W_x|}\sum_{j\in W_x}\bar\ell_j^{(m)}\right]
    -\left[\frac{1}{21}\sum_{j=40}^{60}\bar\ell_j^{(\mathrm{base})}-\frac{1}{|W_x|}\sum_{j\in W_x}\bar\ell_j^{(\mathrm{base})}\right].
    \label{eq:icl-score-position}
\end{equation}
This is the model's early-to-late loss reduction minus the baseline's. Equivalently, $\Delta S_m(x)$ equals the model's $\Delta$NLL over positions 40--60 minus its $\Delta$NLL in the bin ending at $x$, the value plotted in the top row of \cref{fig:appendix-context}. Scores and the full-context NLL references are computed from raw position means.

\FloatBarrier

\subsection{Few-Shot Scaling on Downstream Tasks}
\label{app:fewshot-scaling}

We measure downstream performance as a function of the number of demonstrations, following the demonstration-task protocol also used by SMELT \citep{wang2026smelt}. We use lm-eval-harness version 0.4.3, bf16 weights, and a maximum input length equal to the training context. Results are averaged over three demonstration seeds. We re-tokenize requests to check that prompts fit and cap each task's shot range to avoid truncation.

\begin{figure}[!htbp]
    \centering
    \includegraphics[width=\linewidth]{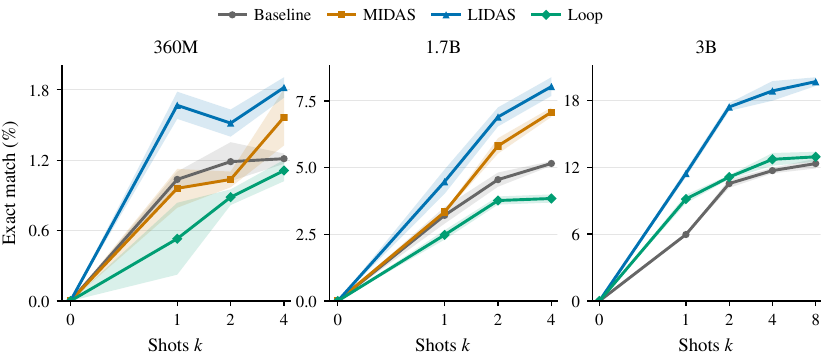}
    \caption{\textbf{Few-shot GSM8K performance across model sizes.} Panels show mean strict exact-match accuracy for 360M, 1.7B, and 3B, with standard-error bands over three demonstration seeds. The shot range is $k\in\{0,1,2,4\}$ for 360M and 1.7B and extends to eight shots for 3B. Details in \cref{app:fewshot-scaling} for the evaluation and loop configurations are specified in \cref{app:loop-configurations}.}
    \label{fig:appendix-gsm8k}
\end{figure}

\paragraph{GSM8K.}
The main comparison in \cref{fig:language-icl}(c) reports GSM8K at 3B for $k\in\{0,1,2,4,8\}$. We score final answers extracted from generated solutions using strict exact match. \cref{fig:appendix-gsm8k} extends this comparison to 360M and 1.7B, using $k\in\{0,1,2,4\}$ for their shorter contexts.

\begin{figure}[!htbp]
    \centering
    \includegraphics[width=\linewidth]{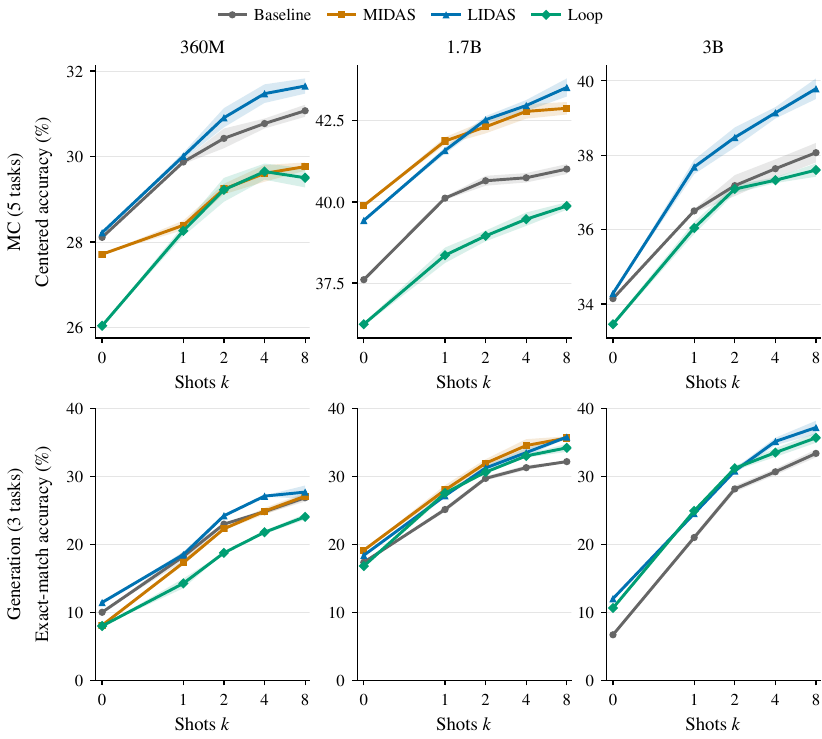}
    \caption{\textbf{Few-shot performance on additional multiple-choice and generation tasks.} Columns show 360M, 1.7B, and 3B. The top row averages chance-centered accuracy over five multiple-choice tasks; the bottom row averages exact-match accuracy over three generation tasks, for $k\in\{0,1,2,4,8\}$. Tasks have equal weight within each subset; bands show standard error across three demonstration-seed subset averages. Individual tasks and normalization are given in \cref{fig:appendix-fewshot-tasks,app:fewshot-scaling}. Loop configurations are specified in \cref{app:loop-configurations}.}
    \label{fig:appendix-fewshot-subsets}
\end{figure}

\paragraph{Additional multiple-choice and generation tasks.}
\cref{fig:appendix-fewshot-subsets} reports separate averages over two fixed task sets, both evaluated at $k\in\{0,1,2,4,8\}$ across all model sizes. The five multiple-choice tasks are ARC-Easy, ARC-Challenge, CommonsenseQA, HellaSwag, and PIQA. ARC, HellaSwag, and PIQA use length-normalized accuracy; CommonsenseQA uses accuracy. The three additional generation tasks are the BigBench tasks QA-Wikidata, Operators, and Repeat-Copy-Logic, scored by exact match after whitespace stripping, ignoring case and punctuation.

\begin{figure}[!htbp]
    \centering
    \includegraphics[width=\linewidth]{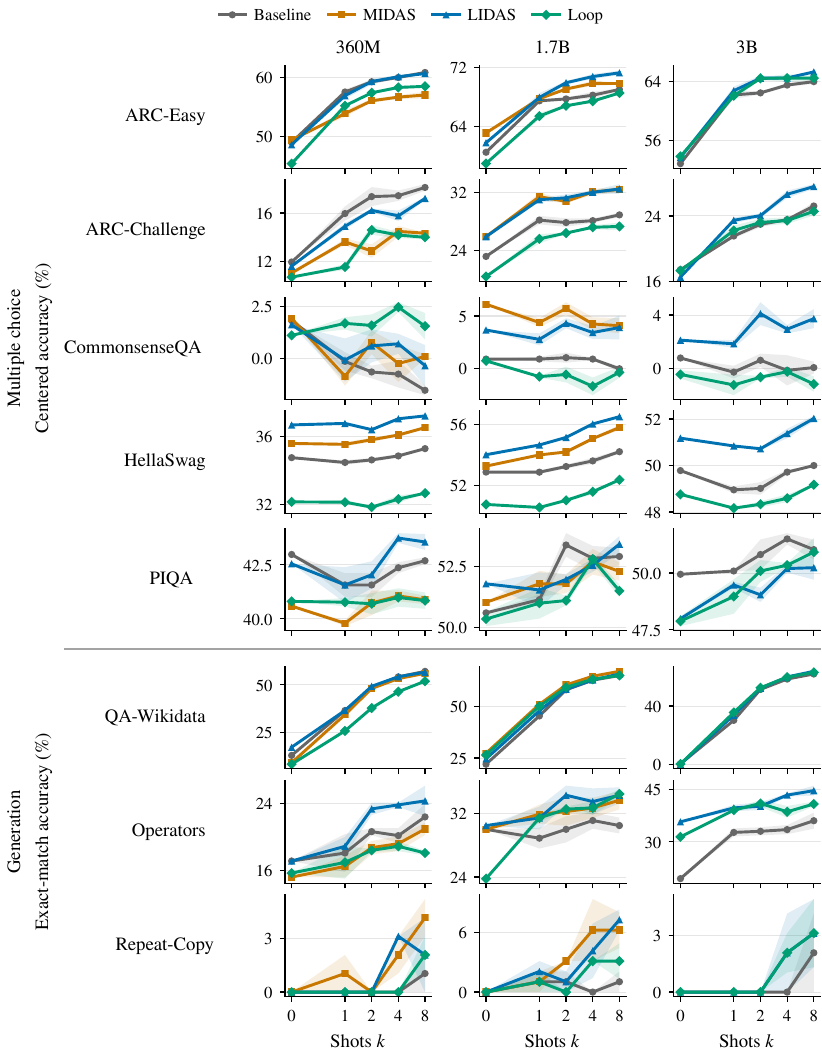}
    \caption{\textbf{Few-shot performance on individual multiple-choice and generation tasks.} Columns show 360M, 1.7B, and 3B. The top five rows show chance-centered accuracy for multiple-choice tasks; the bottom three show exact-match accuracy for generation tasks. Bands show standard error across three demonstration seeds at $k\in\{0,1,2,4,8\}$. Subset averages and normalization are given in \cref{fig:appendix-fewshot-subsets,app:fewshot-scaling}. Loop configurations are specified in \cref{app:loop-configurations}.}
    \label{fig:appendix-fewshot-tasks}
\end{figure}

For the multiple-choice task set $\mathcal{T}$, we report mean chance-centered accuracy:
\begin{equation}
    A_{\mathrm{centered}}(\mathcal{T})=\frac{100}{|\mathcal{T}|}\sum_{t\in\mathcal{T}}\frac{a_t-c_t}{1-c_t}.
    \label{eq:centered-accuracy}
\end{equation}
Here $a_t$ is task accuracy averaged over demonstration seeds, and $c_t$ is $0.25$ for ARC-Easy, ARC-Challenge, and HellaSwag, $0.5$ for PIQA, and $0.2$ for CommonsenseQA. Zero centered accuracy denotes chance performance and 100 denotes perfect performance. For the three generation tasks, we report the equally weighted mean of raw exact-match accuracies in percent, without chance centering. \cref{fig:appendix-fewshot-tasks} shows the individual task curves. Appendix bands show standard error across demonstration seeds, computed from per-seed task averages for the two subsets; they do not estimate uncertainty across training runs.

\FloatBarrier

\subsection{In-Context Graph Tracing}
\label{app:word-grid}

\begin{figure}[!htbp]
    \centering
    \includegraphics[width=\linewidth]{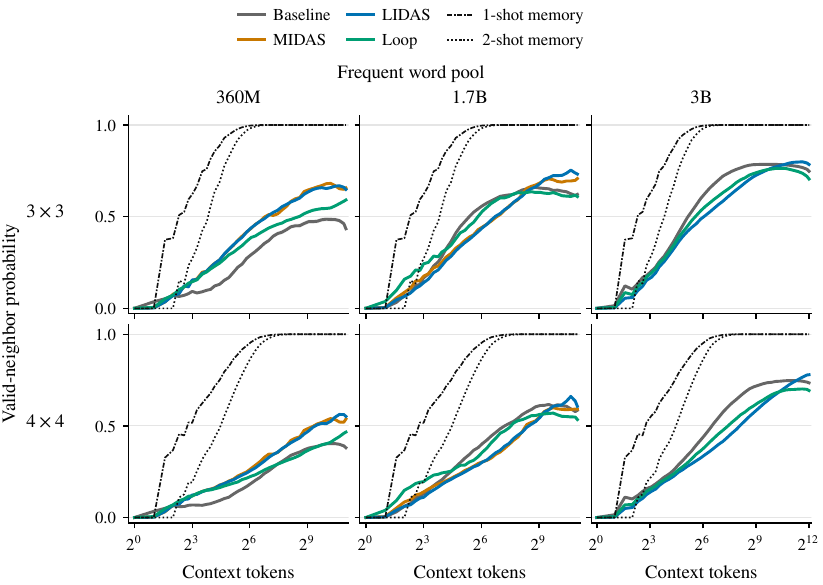}
    \caption{\textbf{Word-grid tracing with frequent-word labels.} Columns show 360M, 1.7B, and 3B; rows show $3\times3$ and $4\times4$ grids. Curves give full-vocabulary probability mass assigned to valid neighbors, with standard error over 200 trials. Black curves are one- and two-observation edge-memorization references. All panels share the probability scale and use base-2 logarithmic context axes. Both settings include the same selected methods at each model size. See \cref{app:word-grid} for the protocol. Loop configurations are specified in \cref{app:loop-configurations}.}
    \label{fig:appendix-word-grid}
\end{figure}

We additionally evaluate the grid-graph setting of \citet{park2025iclr}. Nine or sixteen distinct single-token nouns label the nodes of a $3\times3$ or $4\times4$ grid, respectively. \cref{fig:appendix-word-grid} shows both grid sizes with frequent-word labels for all three model sizes. A random walk is presented as a space-separated sequence of node words, without an additional delimiter. After each node, we sum the model's full-vocabulary next-token probabilities over the words labeling that node's valid neighbors. This probability mass is the graph task's ``accuracy''.

The frequent-word comparison starts from a curated pool of 250 nouns spanning 18 semantic fields. For each tokenizer, we retain nouns that form a single token when preceded by a space and select the 64 with the lowest token IDs. The 360M/1.7B models use cosmo2, whereas 3B uses the SmolLM3 tokenizer; token ID serves as a frequency proxy. Each of 200 trials draws a new node-word assignment and random walks, with one sequence per starting node; scores are averaged over starting nodes within the trial. Models of a given size see the same trials. The horizontal axis counts node tokens in context, up to the native training context length, and bands show the standard error over trials. The one- and two-observation references measure the fraction of positions whose current node has already appeared as a source of a transition at least once or twice. They indicate when copying a previously observed successor becomes possible, rather than upper bounds on graph learning. The 3B comparison includes the dense, \LD, and looped models.

\FloatBarrier

\section{RegBench: Setup and Additional Results}
\label{app:regbench}

\subsection{Task and Data}
\label{app:regbench-data}

We use the released RegBench implementation of \citet{akyurek2024icll}. Each sequence concatenates random walks from one DFA, separated by \texttt{|}. Candidate automata have 4--12 states, an alphabet of 4--18 symbols, and one to three outgoing transitions per state; they are minimized and trimmed before splitting. Each automaton supplies ten to nineteen strings of 1--49 symbols. The fixed dataset contains 40,000 training, 500 validation, and 500 test sequences, split by sampled automaton. All models see the same data. Training and validation cross-entropy mask separator targets and padding.

\subsection{Models, Optimization, and Selection}
\label{app:regbench-training}

We use GPT-2-style pre-LayerNorm Transformers with learned token and position embeddings, two attention heads, a GELU MLP of width $4d$, and tied input/output embeddings. Embedding and attention dropout are 0.1. Dense models have twelve independent layers; the looped model applies two shared layers six times. Grown models reach depth twelve by duplicating the middle $b$ layers and their AdamW moments at each growth event. Copies then train independently, retaining the global learning-rate schedule without restarting warm-up.

All runs use AdamW with $(\beta_1,\beta_2)=(0.9,0.999)$, batch size 32, and 200 epochs (250,000 steps). The learning rate warms up for 25,000 steps and then decays by a cosine schedule to one tenth of its peak. For each configuration, we select the lowest-validation-loss cell over widths $d\in\{64,128,256\}$, peak learning rates $\{10^{-4},2.5\times10^{-4}\}$, and weight decays $\{0.01,0.1\}$. Validation runs after every epoch; training completes the full budget. We evaluate the best-validation checkpoint, restricting grown models to their final full-depth stage, and repeat the selected cells with four training seeds.

\cref{fig:appendix-regbench} includes all five repeated configurations: dense 12L, \LoopTwoSix, and grown $b=2$/prop-1, $b=2$/prop-2, and $b=3$/prop-3. Their widths are 128 except for \LoopTwoSix, which uses 256. All select weight decay 0.1; the peak learning rate is $10^{-4}$ for prop-1 and $2.5\times10^{-4}$ otherwise. These schedules extend the proportional allocation in \cref{app:lm-architectures}, dividing the full training budget among stages in proportion to depth raised to powers 1, 2, and 3, respectively. The main figure uses $b=3$/prop-3, selected by mean validation loss across seeds. Selection never uses test performance. Comparisons match data and optimizer steps, but not width or compute.

\subsection{Metrics and Position Decomposition}
\label{app:regbench-evaluation}

Let $q$ be uniform over the DFA's legal outgoing symbols and zero elsewhere. We report $\operatorname{TVD}(p,q)=\tfrac12\sum_v|p(v)-q(v)|$ and DFA accuracy, which scores whether the argmax is a legal continuation rather than the particular sampled next symbol. Following the released evaluator, TVD includes all non-padding output positions, including separator targets and the final position; $q$ assigns no mass to the separator. Accuracy excludes separator targets and includes the final position. These masks differ from the training loss.

\begin{figure}[!htbp]
    \centering
    \includegraphics[width=\linewidth]{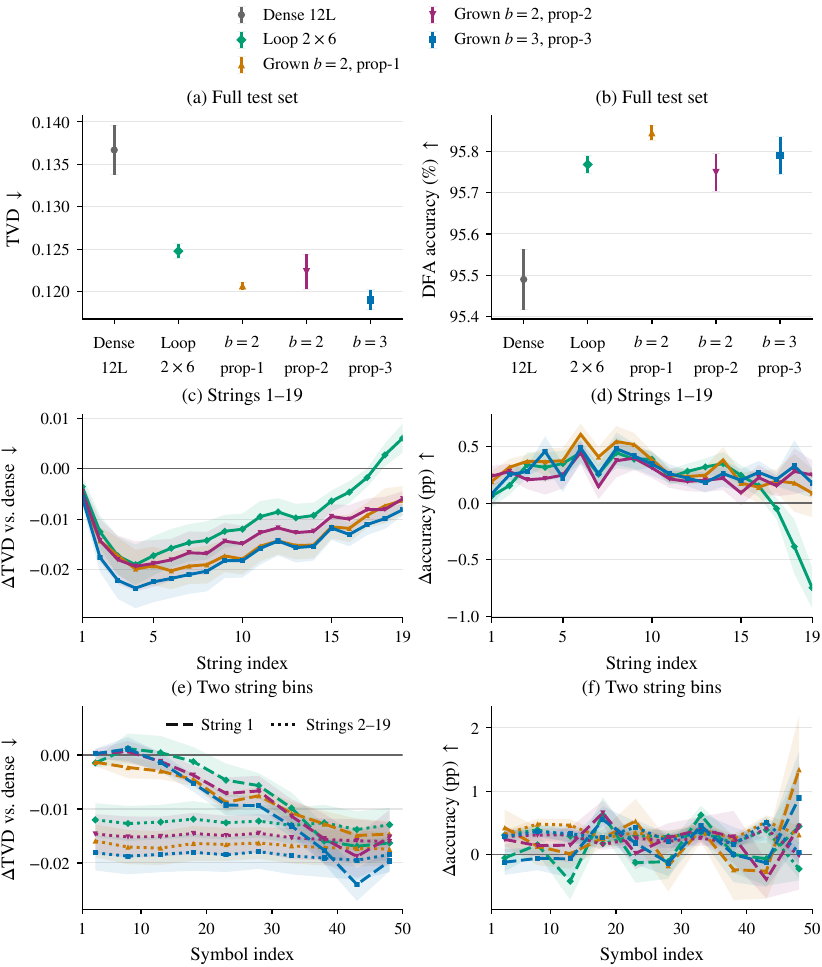}
    \caption{\textbf{RegBench across all configurations evaluated with four seeds.} Columns show TVD (left) and DFA accuracy (right). Rows show full-test means (top), differences from dense by string index (middle), and differences by symbol index (bottom) for string 1 (dashed) and strings 2--19 (dotted). All available strings and symbols are retained; indices start at 1 and symbol positions use bins of five. The positive mean \LoopTwoSix TVD differences at strings 18 and 19 concern only 102 (20.4\%) and 54 (10.8\%) of the 500 test sequences, respectively. Accuracy differences are in percentage points. Error bars and difference bands show standard errors across training seeds. All models have effective depth twelve; validation-selected widths are 128 except for \LoopTwoSix (256). For additional details see \cref{app:regbench-evaluation}.}
    \label{fig:appendix-regbench}
\end{figure}

\cref{fig:regbench-icl,fig:appendix-regbench} report full-test means and position-wise differences from dense. Full-test scores retain the evaluator's batch aggregation; position statistics pool sums and counts within each seed, then average over seeds. String and symbol indices start at 1. In \cref{fig:regbench-icl}, we retain only string indices present in every one of the 500 test sequences (strings 1--10), using all scored symbol positions within these strings (through index 50 for TVD). No symbol-range, coverage, or minimum-support cutoff is applied; only empty bins are omitted. The appendix figure uses all strings 1--19 and all available symbol positions (through 50 for TVD and 49 for accuracy), without range or coverage cutoffs; only empty bins are omitted. Within each figure and metric, the string-index and symbol-index views use the same population; the latter pools symbol indices in bins of five and separates string 1 from the remaining strings (2--10 in the main figure and 2--19 in the appendix).

Error bars show standard error over the four training seeds. Bands on differences use the unpaired standard error $\sqrt{s_{\mathrm{model}}^2/4+s_{\mathrm{dense}}^2/4}$, where $s$ is the sample standard deviation of seed-specific bin means. Negative TVD differences and positive accuracy differences favor the alternative model. Position averages need not equal the full-test statistics because of these different aggregations and, in the main figure, the string restriction.

\subsection{Results Across Configurations}
\label{app:regbench-results}

\FloatBarrier

\section{Omniglot Setup}
\label{app:omniglot}

\subsection{Data, Sequence Format, and Architecture}
\label{app:omniglot-data}

We follow the image-based setup of \citet{chan2022bursty}. We merge the Omniglot training and test splits, giving 1623 character classes with twenty $105\times105$ grayscale exemplars each. A fixed class permutation reserves ten classes for held-out evaluation, while the remaining 1613 classes are sampled uniformly during training. We use no image augmentation, downsampling, or pixel noise. Repeated occurrences of a class within a sequence use distinct exemplars sampled without replacement.

Each sequence contains eight image--label context pairs and one query image. Their embeddings are interleaved as $[x_1,y_1,\ldots,x_8,y_8,x_9]$, giving seventeen tokens, with sinusoidal absolute positional encodings. Training minimizes cross-entropy only at the query position. Images are embedded by a jointly trained small ResNet: a $7\times7$ stride-two convolution with 64 channels, batch normalization, ReLU, and a $3\times3$ stride-two max pool, followed by four groups of two bottleneck residual blocks with output channels 16/32/32/64 and strides 1/2/2/2, and global average pooling. Labels use a learned $1623\times64$ embedding table.

The sequence model is a causal pre-LayerNorm Transformer of width 64, with eight attention heads, GELU MLPs of width 256, no dropout, a final LayerNorm, and a linear head with 1623 outputs. \cref{fig:omniglot-icl} compares dense depths 4 and 12, a \LoopFourThree model, and a \LD{}-grown model. The latter branches from the four-layer dense run, grows to eight layers at step 50,000, and to twelve at step 100,000 by duplicating the middle four layers. We copy Adam moments together with weights and then train all copies independently, retaining the global step and learning-rate schedule. Thus, the grown model and the four-layer reference have exactly the same pre-growth trajectory. The looped model shares only Transformer blocks, the image encoder, embeddings, and final readout are not repeated and are identical in setup for all versions.

\subsection{Training Distributions and Optimization}
\label{app:omniglot-training}

\paragraph{Competing context and weight solutions.}
In \cref{fig:omniglot-icl}(a), each training class has a fixed label and every sequence is bursty, $p(\mathrm{bursty})=1$. We sample two training classes, place three exemplars of each in the context, add two further uniformly sampled training items, and shuffle the eight context items. The query is a new exemplar of one of the two burst classes. Its label can be recovered either by matching it to context examples or by recalling its fixed training-class association.

\paragraph{ICL-only training.}
In \cref{fig:omniglot-icl}(b), training uses four-shot, two-way sequences: two training classes each contribute four context exemplars, and the query is a fifth exemplar of one of them. Labels 0 and 1 are reassigned randomly to the two classes in every sequence. There is therefore no persistent class-to-answer mapping that can solve the query without context. We intentionally only modify the training data, i.e., the model still has 1623 output logits, although only labels 0 and 1 are correct in this setting. Architecture, optimizer, and sequence length also remain unchanged.

Both settings use Adam with $(\beta_1,\beta_2)=(0.9,0.999)$ and $\epsilon=10^{-8}$, batch size 32, no weight decay or gradient clipping, and learning rate
\begin{equation}
    \eta(t)=3\times10^{-4}\min\left(\frac{t}{4000},\sqrt{\frac{4000}{t}}\right),\qquad t>0.
\end{equation}
This gives 4000 steps of linear warm-up followed by inverse-square-root decay. Each fixed-label configuration uses four seeds, and each ICL-only configuration uses eight. These control initialization and sampling independently of the fixed class split. For the reported comparisons, we use 40M training steps in the fixed-label setting and 1M training steps in the ICL-only setting. Checkpoints are saved every 50,000 steps and at powers of two starting at 64 for finer early-training resolution.

\subsection{Evaluation: ICL and IWL}
\label{app:omniglot-evaluation}

The ICL probe uses four-shot, two-way sequences from the ten held-out classes, with new class-to-label assignments in each sequence. The context contains four distinct exemplars per class, and the query is a fifth exemplar of one of them. We report accuracy with the argmax restricted to logits 0 and 1 (chance $1/2$). Success therefore requires using the current context for a class never encountered in training. The IWL probe instead samples the query uniformly from the 1613 training classes, uses its fixed training label, and guarantees that its class is absent from the eight context examples. It reports the unrestricted argmax over all 1623 outputs, consequently, uniform guessing over that output space gives chance $1/1623$. The IWL probe is used only for the fixed-label setting.

Every checkpoint is evaluated on the same sequences generated with a fixed evaluation seed. The fixed-label setting uses 10,000 sequences per probe and the ICL-only setting uses 2000 for its ICL probe. Curves average over four training seeds in the fixed-label setting and eight in the ICL-only setting, at checkpoints shared by all seeds within each configuration; bands are one standard error across seeds. For display, each seed trajectory is smoothed with a centered, equally weighted moving average over 51 checkpoints in the fixed-label setting and 3 in the ICL-only setting, shortening windows at the boundaries after truncating to the reported training budget. Means and standard errors are then computed across the smoothed seed trajectories. Endpoint statistics and emergence thresholds in the text use unsmoothed evaluations. The two plots in panel (a) use the matched 40M-step horizon, while panel (b) uses the shared 1M-step horizon.

\FloatBarrier

\section{Induction-Head Development and Readouts}
\label{app:induction}

\subsection{Measures and Aggregate Training Trajectories}
\label{app:induction-measures}

We implement the head diagnostics of \citet{yin2025which} and \citet{aoyama2026predicting} on converted language-model checkpoints, using fp32 weights and TF32 matrix multiplications. The same prompts and random sequences are reused across checkpoints and models within each size. Attention-head outputs are measured immediately before the attention output projection. For looped models, each application of a shared head is evaluated separately, since its input and causal role can differ across passes.

\paragraph{Prefix matching.}
We generate 1000 sequences of the form $[\mathrm{doc\mbox{-}start},x_1,\ldots,x_{50},x_1,\ldots,x_{50}]$, drawing 50 distinct regular vocabulary tokens per sequence. For a second-half token, the induction target is the token following its earlier occurrence. Prefix-matching score (PS) is the mean attention probability assigned to the target position, averaged over second-half positions and sequences. With zero-based sequence positions $p\in\{51,\ldots,100\}$, the target position is $p-49$; the last position uses the cyclic continuation $x_1$.

\paragraph{Function-vector scores.}
Following \citet{todd2024function,yin2025which}, we estimate a function-vector (FV) score on 29 training tasks using ten demonstrations and 100 clean and 100 corrupted prompts per task. We do not filter tasks or prompts by performance on clean prompts. For each head and task, we compute its mean last-token output on clean prompts. We then shuffle the demonstration labels and replace that head's last-token output with the clean task mean. The FV score is the patched minus unpatched probability of the correct \emph{first answer token} on the corrupted prompt, averaged over prompts and then tasks. Head selection therefore uses a causal intervention on demonstration-based task execution, rather than the attention pattern alone.

\begin{figure}[!htbp]
    \centering
    \includegraphics[width=\linewidth]{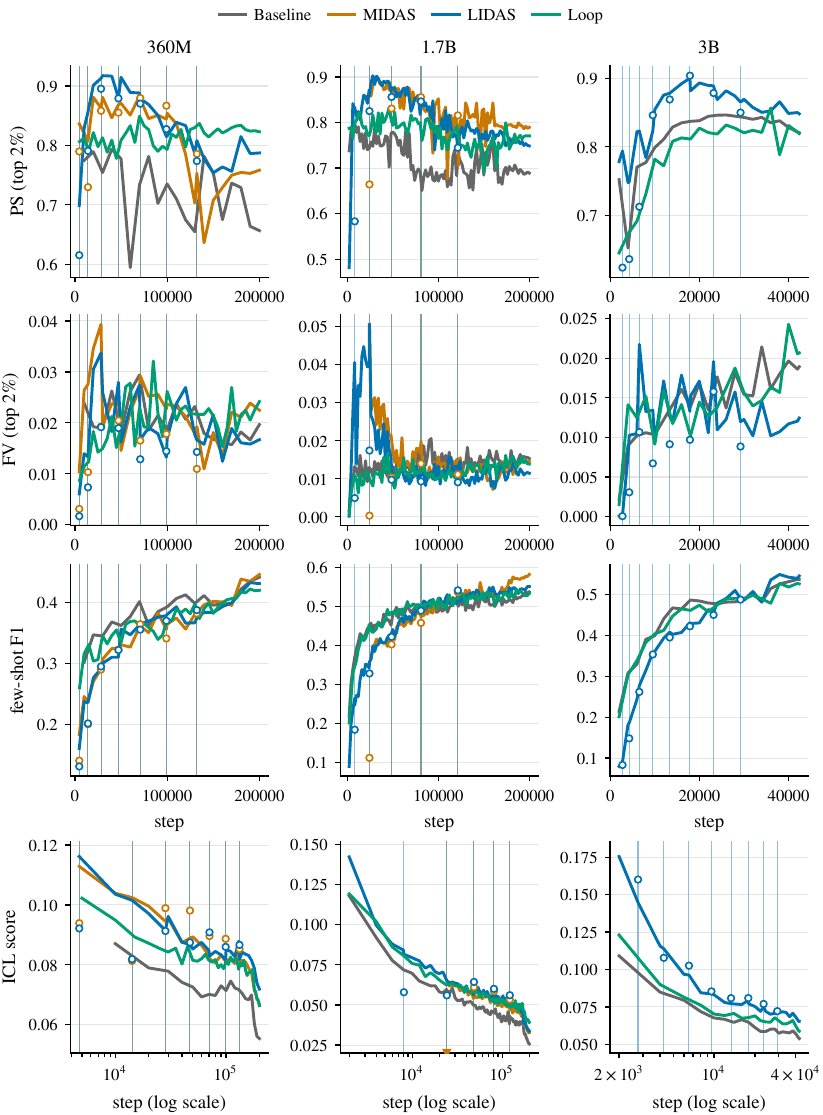}
    \caption{\textbf{Training trajectories across model sizes.} Columns show 360M, 1.7B, and 3B. Rows report PS and FV averaged over their respective checkpoint-specific top-2\% head sets, nineteen-task ten-shot F1, and the window-averaged ICL-Score (\cref{eq:icl-score}). Curves connect raw trained checkpoints, with regular spacing of 10,000 steps at 360M and 2000 steps at 1.7B and 3B, plus additional growth points. At 3B, all models use the shared regular grid. The first three rows use linear steps; the last uses logarithmic steps. Vertical rules mark growth and hollow markers show immediate post-growth evaluations. The downward marker in the 1.7B ICL-Score panel denotes \MD at $-0.723$, below the plotted range. Loop configurations are specified in \cref{app:loop-configurations}.}
    \label{fig:appendix-trajectories}
\end{figure}

\paragraph{Few-shot task performance.}
\cref{fig:head-ablation-combined}, \cref{fig:appendix-trajectories} (third row) and \cref{fig:appendix-readouts} use a separate, fixed ten-shot evaluation from the downstream shot-scaling experiment. It comprises eleven held-out tasks from \citet{todd2024function} and eight additional tasks from the \citet{yin2025which} evaluation: AG News, Capitalize First Letter, CoNLL Organization, National Parks, Park--Country, Person--Occupation, Animal versus Object (three items), Choose First of Five, English--German, Person--Instrument, CommonsenseQA, Capital Index, Capitalize Second Letter, Abstract Classification, French--English, and the Fruit, Capital-Par, Capital, and Shape binding tasks. These tasks are disjoint from those used to estimate FV scores.

Prompts interleave \texttt{Q:} inputs and \texttt{A:} outputs, separate demonstrations by a blank line, and end at the query's \texttt{A:}. A document-start token is prepended. We draw ten demonstrations from each task's demonstration split and evaluate 100 queries, generating up to five tokens greedily. The answer is extracted from the generated continuation, lowercased, stripped of punctuation and English articles, and split on whitespace. We report token-overlap F1 against the reference, averaged first over queries and then over the nineteen tasks. Prompt sampling is fixed per task across checkpoints. This F1 is distinct from the benchmark accuracies in \cref{app:fewshot-scaling}.

\paragraph{Checkpoint-specific summaries.}
In \cref{fig:language-emergence,fig:appendix-trajectories}, PS and FV, respectively, are summarized by averaging the top $\lceil0.02H\rceil$ heads under each measure at each checkpoint, where $H$ is the current number of head applications. At full depth, $H=480$, 768, and 576 for 360M, 1.7B, and 3B, giving 10, 16, and 12 selected heads. These sets are reselected at every checkpoint and may have different membership.

The PS trajectories in \cref{fig:language-emergence} use only trained checkpoints shared by all displayed models, every 5k steps for 360M and 2k steps for 1.7B and 3B. Display smoothing averages up to three adjacent shared checkpoints within 2000 steps of the center, including across growth boundaries. The first and final checkpoints remain raw. Immediate post-growth evaluations are shown only as raw hollow markers and never enter smoothing. In \cref{fig:appendix-trajectories}, the 1.7B \MD post-growth ICL score at step 24,285 is approximately $-0.72$; its marker is placed on the lower boundary because it lies outside the displayed range. The regular checkpoint spacing of the dense and grown main runs does not resolve the earliest induction-head transition.

\paragraph{ICL-related training trajectories.}
\cref{fig:appendix-trajectories} extends the main-paper PS comparison to FV scores, few-shot F1 on nineteen held-out tasks, and the token-loss ICL-Score, across all three model sizes. In the 3B comparison, early growth events disrupt both PS and FV scores, followed by recovery during further training. The grown model reaches higher peak scores than the baseline, then finishes with higher PS but lower FV scores. The looped model finishes close to the baseline in PS and slightly above it in FV score. Few-shot F1 improves during training, with the grown model initially trailing and ultimately exceeding the baseline, while the looped model finishes close to it. Meanwhile, the ICL-Score declines for all models as early-token predictions improve more than later-token predictions. Both grown and looped models nevertheless finish with higher ICL-Scores than the baseline.

\FloatBarrier
\subsection{Final-Head Identities, Lineages, and Locations}
\label{app:head-lineages}

\paragraph{Selection and ancestry.}
We select the final top-2\% heads independently by PS and FV (10, 16, and 12 heads at 360M, 1.7B, and 3B) and trace these fixed identities backward through training. For grown models, we invert the actual layer-duplication maps: a head and its later copies share an ancestral trajectory until duplication, drawn only once. The 1.7B \MD branch uses its shared \LD trunk before step 24,285. Looped layers appear once per application. No particular seed-head pair is assumed to recur across sizes.

\paragraph{Checkpoint selection and smoothing.}
Trajectories begin at step 2000 for 1.7B and 3B; at 360M, they start at the first measured checkpoints: 10,000 for the baseline, 5000 for Loop, and 4722 for the grown models. Exact growth-step measurements are omitted except for this first 360M pre-growth checkpoint, the only measurement of the four-layer source stage. Immediate post-growth scores do not enter the curves. For each head, we apply a triangular weighted mean within each growth stage: at step $t$, a measurement at $t_j$ receives weight proportional to $1-|\log_2(t_j/t)|/0.125$ when $|\log_2(t_j/t)|<0.125$, and zero otherwise. The first and last available measurements of each stage, including final scores, remain raw; smoothing never crosses a growth boundary.

\begin{figure}[!htbp]
    \centering
    \includegraphics[width=\linewidth]{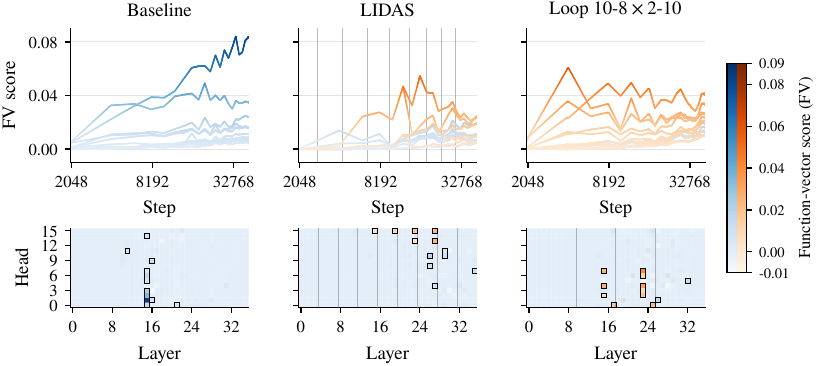}
    \caption{\textbf{FV heads recur at the grown or looped block size but emerge later than induction heads.} FV counterpart of \cref{fig:head-trajectories-locations} for the 3B models. \textbf{Top:} trajectories of the twelve final top-2\% FV heads per model, selected independently of PS. In \LD, these heads trace to eight ancestors; two, L3H13 and L3H15, account for six final heads (two and four, respectively; orange). \textbf{Bottom:} final FV scores for all heads; black outlines mark the selected set. Orange in \LoopThree marks selected heads in the repeated block; other heads are blue. Scores share a common scale that retains negative values. Axes, growth and block guides, lineage tracing, and smoothing follow \cref{fig:head-trajectories-locations} and this section.}
    \label{fig:appendix-fv-trajectories-locations}
\end{figure}

\begin{figure}[!htbp]
    \centering
    \includegraphics[width=\linewidth]{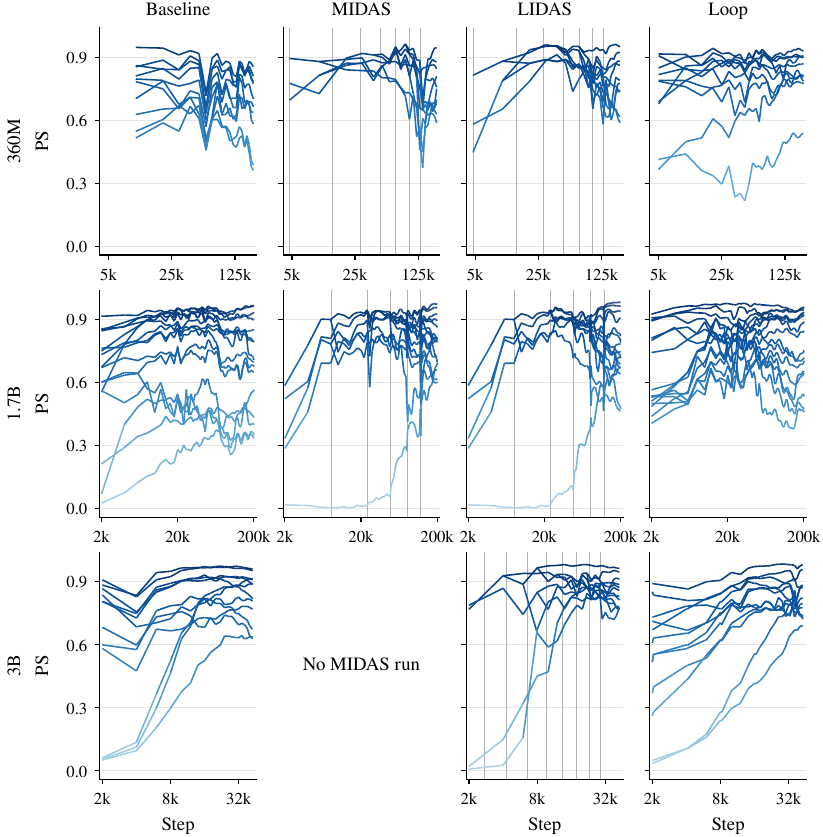}
    \caption{\textbf{Trajectories of the final top-2\% prefix-matching heads.} Rows show 360M, 1.7B, and 3B; columns show Baseline, \MD, \LD, and Loop. Each panel follows the final 10, 16, or 12 heads ranked by PS, respectively, through their actual growth ancestry; sets are not reselected at intermediate checkpoints. Shared ancestral paths are drawn once, and vertical rules mark growth. Lines use the within-stage smoothing and boundary policy in \cref{app:head-lineages}; step ticks marked k denote thousands. Scores share a common vertical range. There is no 3B \MD run. Loop configurations are specified in \cref{app:loop-configurations}.}
    \label{fig:appendix-head-trajectories-ps}
\end{figure}

\begin{figure}[!htbp]
    \centering
    \includegraphics[width=\linewidth]{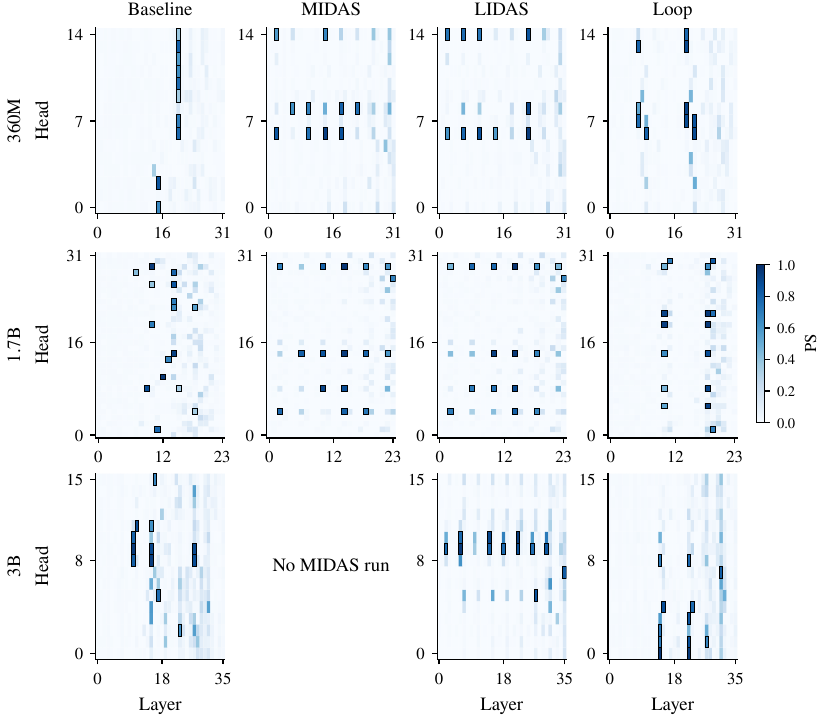}
    \caption{\textbf{Final prefix-matching scores for every attention head.} Rows show 360M, 1.7B, and 3B; columns show Baseline, \MD, \LD, and Loop. All panels share one PS color scale. Black outlines identify the final top-2\% heads traced in \cref{fig:appendix-head-trajectories-ps}. Layer and head indices are zero-based; loop applications occupy separate layer positions. The empty 3B \MD slot denotes an unavailable model. Loop configurations are specified in \cref{app:loop-configurations}.}
    \label{fig:appendix-head-locations}
\end{figure}

\begin{figure}[!htbp]
    \centering
    \includegraphics[width=\linewidth]{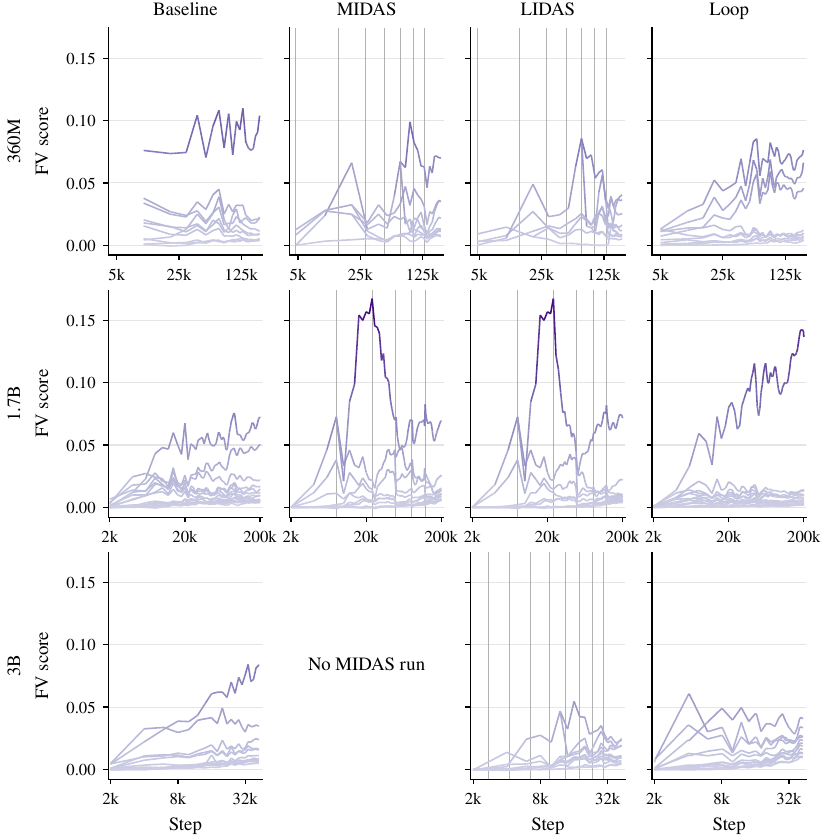}
    \caption{\textbf{Trajectories of the final top-2\% function-vector heads.} Rows show 360M, 1.7B, and 3B; columns show Baseline, \MD, \LD, and Loop. Each panel follows the final 10, 16, or 12 heads ranked by FV, respectively, through their actual growth ancestry; sets are not reselected at intermediate checkpoints. Shared ancestral paths are drawn once, and vertical rules mark growth. Lines use the within-stage smoothing and boundary policy in \cref{app:head-lineages}; step ticks marked k denote thousands. Scores share a common vertical range. There is no 3B \MD run. Loop configurations are specified in \cref{app:loop-configurations}.}
    \label{fig:appendix-head-trajectories-fv}
\end{figure}

\begin{figure}[!htbp]
    \centering
    \includegraphics[width=\linewidth]{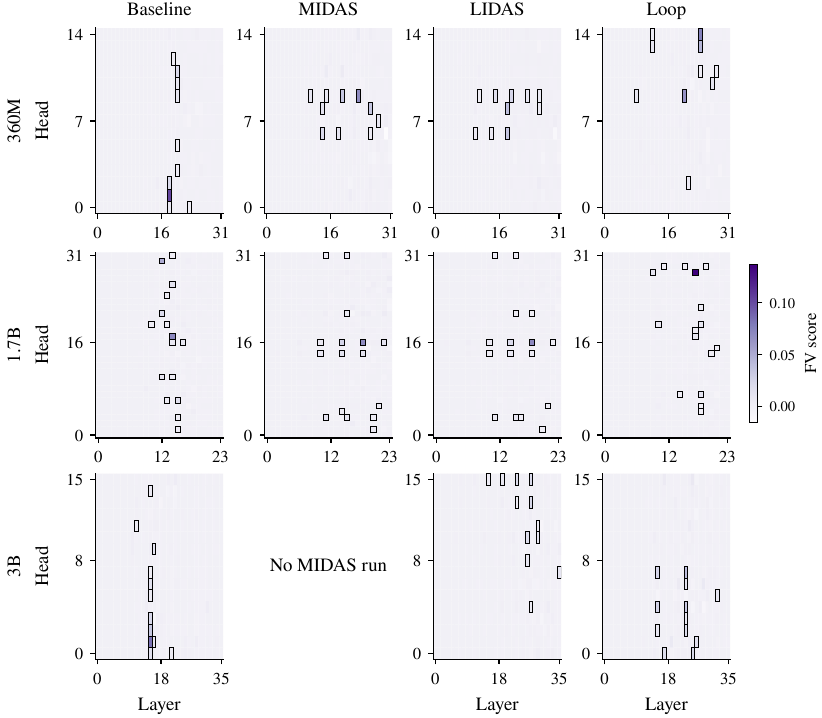}
    \caption{\textbf{Final function-vector scores for every attention head.} Rows show 360M, 1.7B, and 3B; columns show Baseline, \MD, \LD, and Loop. All panels share one FV color scale. Black outlines identify the final top-2\% heads traced in \cref{fig:appendix-head-trajectories-fv}. Layer and head indices are zero-based; loop applications occupy separate layer positions. Negative FV scores, where present, are retained. The empty 3B \MD slot denotes an unavailable model. Loop configurations are specified in \cref{app:loop-configurations}.}
    \label{fig:appendix-head-locations-fv}
\end{figure}

\paragraph{Display conventions.}
\cref{fig:head-trajectories-locations,fig:appendix-fv-trajectories-locations} pair the 3B PS and FV lineages with final all-head score maps. Cross-size trajectories appear in \cref{fig:appendix-head-trajectories-ps,fig:appendix-head-trajectories-fv}, with the selected sets outlined in the corresponding maps (\cref{fig:appendix-head-locations,fig:appendix-head-locations-fv}). Curves are opaque, and color intensity follows the local smoothed score. Each figure uses a common score scale across models and sizes: 0--1 for PS, and a range retaining negative values for FV. Color does not encode final-head rank or PS/FV set overlap. Neither a high score nor repeated locations alone establish a head's causal importance.

\paragraph{Cross-size PS and FV lineages.}
\cref{fig:appendix-head-trajectories-ps,fig:appendix-head-trajectories-fv} independently select the final top-2\% heads by PS and by FV, giving 10, 16, and 12 heads at 360M, 1.7B, and 3B. Each set is then traced by identity through the same inverse duplication maps. The corresponding all-head final score matrices appear in \cref{fig:appendix-head-locations,fig:appendix-head-locations-fv}; outlines mark exactly the heads selected for each trajectory plot. The 1.7B \MD branch uses its shared \LD trunk before step 24,285.

The appendix applies the same within-stage triangular log-step smoother with halfwidth 0.125, preserving stage endpoints and final values. Exact growth-step measurements are omitted except for the first 360M pre-growth checkpoint at step 4722: its explicit pre-growth stage identifies the four-layer source, and it is the only measured point in that stage. Immediate post-growth scores do not enter these lineage curves. The 360M dense and looped curves start at their first measured checkpoints, 10,000 and 5000. Shared ancestral segments are drawn once. PS uses a common 0--1 range; FV uses a common range across models and sizes that includes all displayed values. Color intensity follows the local score and does not encode model identity or PS/FV set overlap. No particular seed-head pair is assumed to recur across sizes.

\FloatBarrier

\FloatBarrier

\subsection{Per-Layer Readouts}
\label{app:layer-readout}

\begin{figure}[!htbp]
    \centering
    \includegraphics[width=\linewidth]{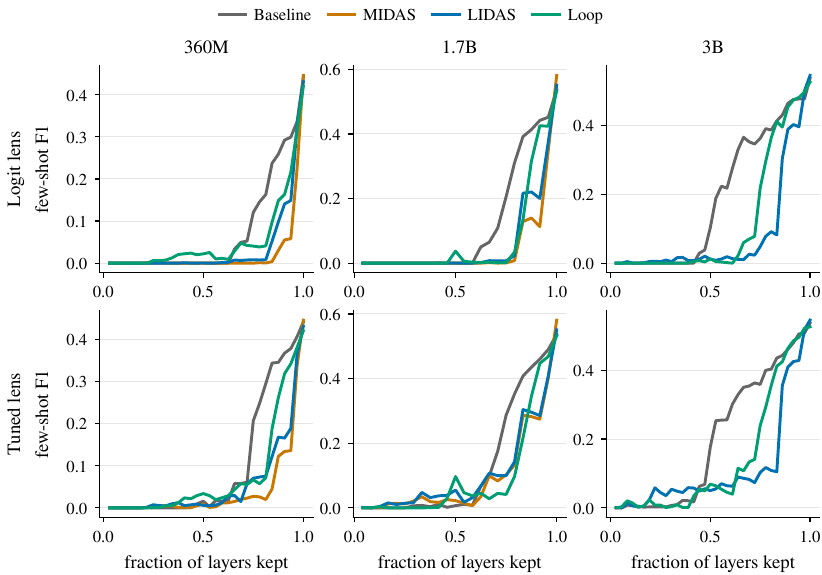}
    \caption{\textbf{Intermediate few-shot readouts across model sizes.} Columns show 360M, 1.7B, and 3B; rows use the logit and tuned lenses. Each curve runs the nineteen-task, ten-shot F1 evaluation after the indicated fraction of layers. The tuned lens is trained separately for each frozen language model by next-token distillation (\cref{app:layer-readout}). These readouts measure accessibility of predictions, not the causal importance of a layer. Loop configurations are specified in \cref{app:loop-configurations}.}
    \label{fig:appendix-readouts}
\end{figure}

\cref{fig:appendix-readouts} evaluates intermediate predictions at each exit depth $l$, with the horizontal axis normalized to $l/L$. For each generated token, we run only the first $l$ layers and decode their residual stream. The logit lens applies the full model's final normalization and unembedding directly. The tuned lens \citep{belrose2023eliciting} first applies a learned affine residual translator $h\mapsto h+W_lh+b_l$. Translators are initialized at zero, so the initial lens is the identity, and the language model remains frozen.

Each model's translators are trained on its own FineWeb-Edu pretraining source for 250 optimizer steps with $2^{18}$ tokens per step, minimizing KL divergence from the full model's next-token distribution to the translated intermediate prediction. Training uses SGD with Nesterov momentum 0.9, initial learning rate 0.1 with linear decay, weight decay $10^{-3}$, gradient clipping at norm 1, and bf16 autocast. Tokenization and document masking follow the corresponding pretraining setup. Translator index $l$ decodes the residual entering layer $l$ (equivalently, after $l$ completed layers); the full-depth exit uses the original readout without a translator.

We run the same nineteen-task, ten-shot F1 evaluation at each exit depth. An identity-translator (zero-lens) control reproduces the logit lens, and the full-depth exit reproduces the original model. These probes test how accessible predictions are to a fixed or learned readout; they do not by themselves identify which layers causally compute the prediction.

\FloatBarrier

\section{Attention-Head Ablations}
\label{app:head-ablation}

\subsection{Head Selection and Mean Ablation}
\label{app:head-ablation-protocol}

We mean-ablate fractions $f\in\{0.01,0.03,0.05,0.07,0.09,0.15,0.20\}$ of final-checkpoint head applications. The combined figure uses exclusive PS and FV rankings; additional sweeps also include non-exclusive and random rankings. Exclusive induction rankings first exclude the top $\lfloor0.02H\rfloor$ FV heads; exclusive FV rankings analogously exclude the top induction heads. The number ablated remains $\lfloor fH\rfloor$, measured relative to the full set before exclusion. Joint rankings use the geometric mean of PS and FV after clipping each score at zero and dividing by its maximum. A selected head's output slice is replaced by a mean activation immediately before the attention output projection. In looped models, selection and intervention act on individual applications, so the ablated fraction is not a fraction of unique parameters.

\subsection{Measurements and Plotted Summaries}
\label{app:ablation-windowed}
\label{app:ablation-position}
\label{app:ablation-additional}

\begin{figure}[!htbp]
    \centering
    \includegraphics[width=\linewidth]{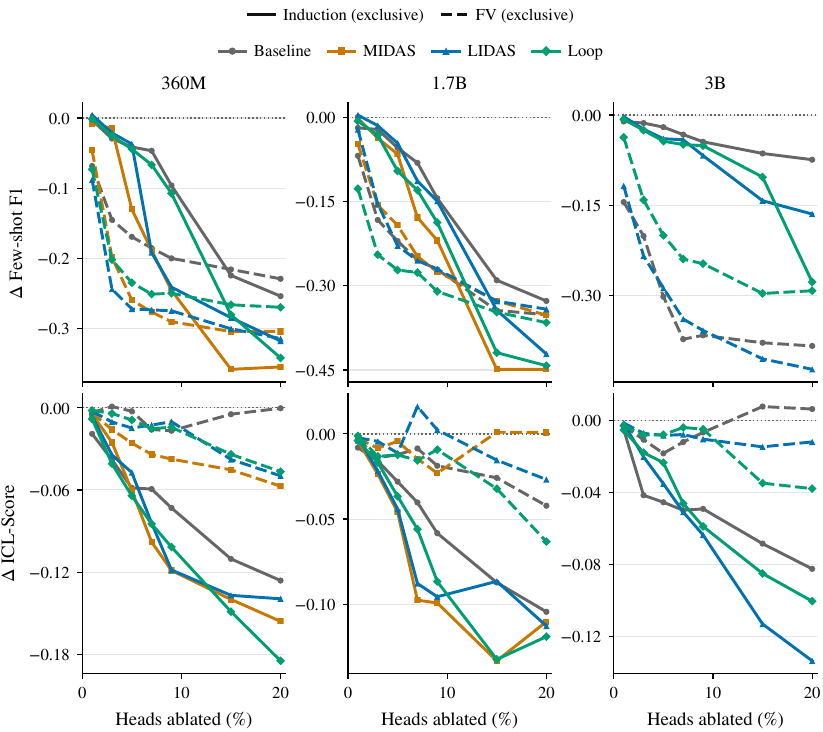}
    \caption{\textbf{Exclusive head ablations across model sizes.} Columns show 360M, 1.7B, and 3B. Top: change in nineteen-task ten-shot F1. Bottom: change in windowed ICL-Score (\cref{eq:ablation-windowed-delta}). Solid lines ablate induction-ranked heads after excluding the top-2\% FV heads; dashed lines reverse these roles. Markers distinguish models. Fractions are 1\%, 3\%, 5\%, 7\%, 9\%, 15\%, and 20\%; zero denotes the corresponding unablated result. Loop configurations are specified in \cref{app:loop-configurations}.}
    \label{fig:appendix-exclusive-ablation}
\end{figure}

\begin{figure}[!htbp]
    \centering
    \includegraphics[width=\linewidth]{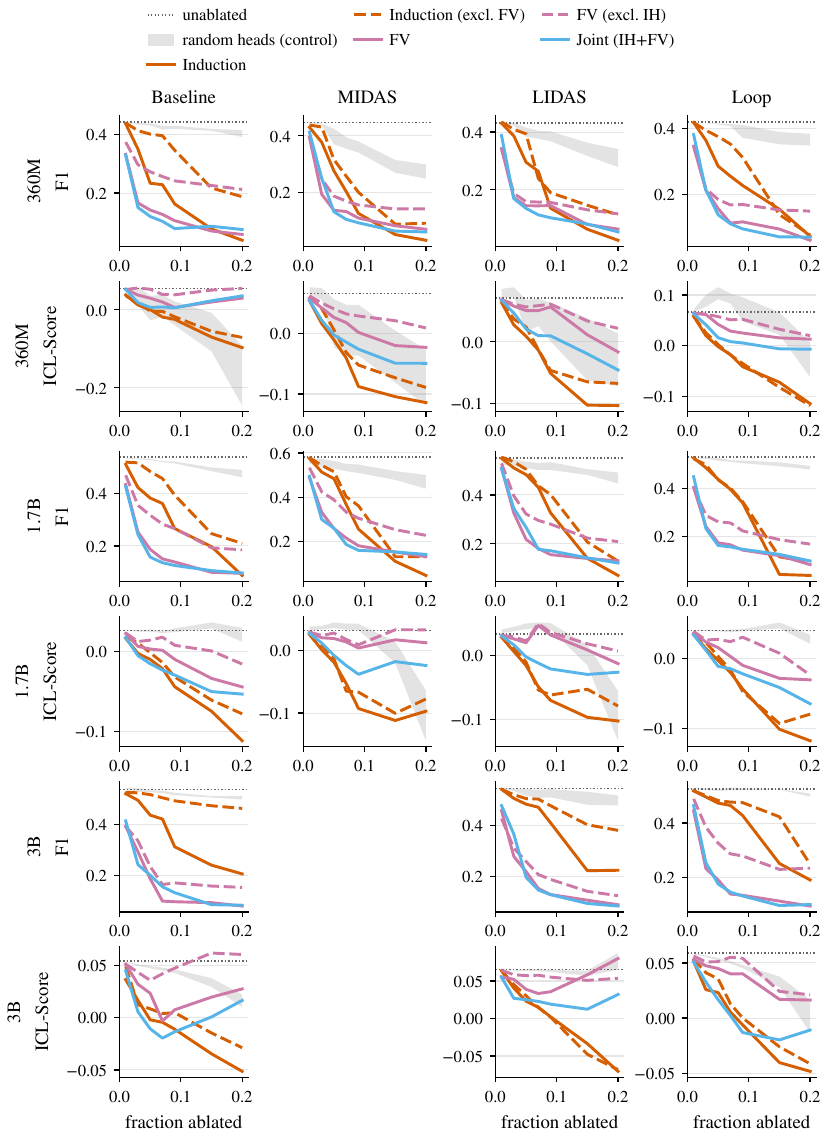}
    \caption{\textbf{Head ablation across model sizes.} Row pairs show 360M, 1.7B, and 3B, each with absolute few-shot F1 above the windowed ICL-Score (\cref{eq:icl-score}); columns show dense, \MD, \LD, and looped models. Dashed rankings exclude the other measure's top-2\% heads. Gray bands show the mean $\pm$ standard error over random head sets: three for F1 and sixteen for the ICL-Score at every model size. Dotted lines give unablated values. Fractions count applications in looped models. Panel scales vary; the empty 3B \MD column reflects the missing model. See \cref{app:head-ablation}. Loop configurations are specified in \cref{app:loop-configurations}.}
    \label{fig:appendix-ablation}
\end{figure}

\paragraph{Few-shot F1.}
The replacement is the clean last-token head activation averaged over the 29 FV training tasks. It is applied at the last position of the prompt and each subsequent generation forward pass. We repeat the nineteen-task ten-shot F1 evaluation of \cref{app:induction-measures}. \cref{fig:head-ablation-combined}(a) and \cref{fig:appendix-exclusive-ablation} report changes from each model's unablated F1, while \cref{fig:appendix-ablation} reports absolute F1.

\paragraph{Token loss.}
We measure token-level NLL on 10,000 FineWeb-Edu documents, replacing selected heads' outputs with their natural-text mean activations at all positions. Let $\bar\ell_j$ be the document-averaged NLL at prediction position $j$ (\cref{eq:position-nll}) and $\delta_j=\bar\ell_j^{(\mathrm{ablated})}-\bar\ell_j^{(\mathrm{clean})}$ its increase from the same model without ablation on the same documents. Position-wise plots report $\delta_j$. We also summarize token loss using the windowed ICL-Score in \cref{eq:icl-score}; \cref{fig:head-ablation-combined}(b) and \cref{fig:appendix-exclusive-ablation} report its change:
\begin{equation}
    \Delta S_{\mathrm{ICL}}=S_{\mathrm{ICL}}^{(\mathrm{ablated})}-S_{\mathrm{ICL}}^{(\mathrm{clean})}.
    \label{eq:ablation-windowed-delta}
\end{equation}

\cref{fig:appendix-ablation} reports the absolute ICL-Score.

Position curves average $\delta_j$ in logarithmic bins (24 per decade) for display on a base-2 axis. In \cref{fig:head-ablation-combined}(c), all six curves additionally receive Gaussian-weighted local linear smoothing in $\log_2$ position with bandwidth $\sigma=0.2$. Windowed scores use raw position losses before any binning or display smoothing.

\begin{figure}[!htbp]
    \centering
    \includegraphics[width=\linewidth]{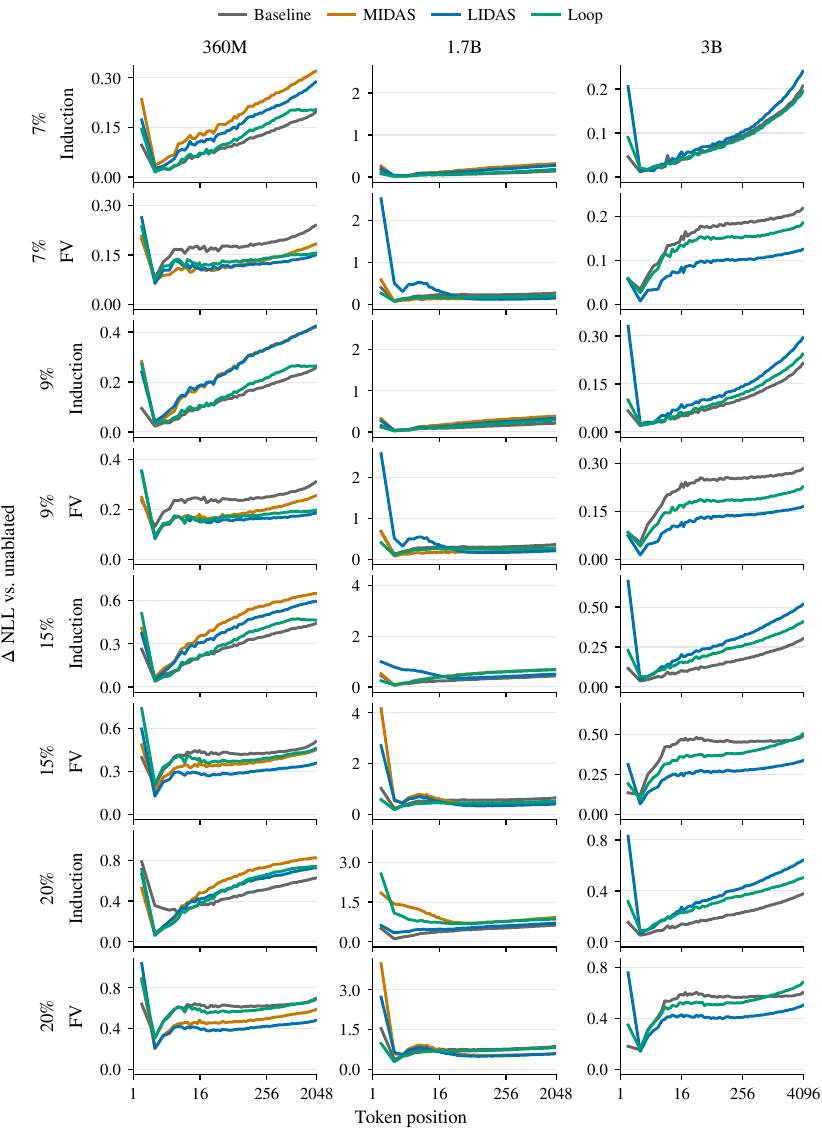}
    \caption{\textbf{Position-wise head ablation at higher fractions.} Columns show 360M, 1.7B, and 3B; row pairs show exclusive induction and FV ablation at 7\%, 9\%, 15\%, and 20\%. Curves give the mean NLL increase on 10,000 FineWeb-Edu documents, with log-binned positions on a base-2 axis. Each induction/FV pair within a size shares an uncropped vertical range. \MD is unavailable at 3B. See \cref{app:head-ablation}. Loop configurations are specified in \cref{app:loop-configurations}.}
    \label{fig:appendix-ablation-position}
\end{figure}

\end{document}